\documentclass[letterpaper]{article} % DO NOT CHANGE THIS
\usepackage[]{aaai25}  % DO NOT CHANGE THIS
\usepackage{times}  % DO NOT CHANGE THIS
\usepackage{helvet}  % DO NOT CHANGE THIS
\usepackage{courier}  % DO NOT CHANGE THIS
\usepackage[hyphens]{url}  % DO NOT CHANGE THIS
\usepackage{graphicx} % DO NOT CHANGE THIS
\usepackage{natbib}  % DO NOT CHANGE THIS AND DO NOT ADD ANY OPTIONS TO IT
\usepackage{caption} % DO NOT CHANGE THIS AND DO NOT ADD ANY OPTIONS TO IT
\usepackage{multicol}
\usepackage[utf8]{inputenc} % allow utf-8 input
\usepackage[T1]{fontenc}    % use 8-bit T1 fonts
\usepackage{url}            % simple URL typesetting
\usepackage{booktabs}       % professional-quality tables
\usepackage{amsfonts}       % blackboard math symbols
\usepackage{nicefrac}       % compact symbols for 1/2, etc.
\usepackage{microtype}      % microtypography
\usepackage{xcolor}         % colors
\usepackage{algorithm}
\usepackage{algorithmic}
\usepackage{graphicx}
\usepackage{enumitem}
\usepackage{dsfont}
\usepackage{amsmath}
\usepackage{multirow}
\usepackage{amsmath}
\usepackage{amssymb}
\usepackage{mathtools}
\usepackage{amsthm}

\usepackage{lipsum}
\usepackage{multicol}

\usepackage[T1]{fontenc}
\usepackage{array}
\usepackage{booktabs}
\usepackage{subcaption}

\definecolor{cadetblue}{rgb}{0.37, 0.62, 0.63}
\definecolor{caribbeangreen}{rgb}{0.0, 0.8, 0.6}
\definecolor{darkspringgreen}{rgb}{0.13, 0.55, 0.13}
\definecolor{frenchrose}{rgb}{0.96, 0.29, 0.54}

\usepackage{newfloat}
\usepackage{listings}
\DeclareCaptionStyle{ruled}{labelfont=normalfont,labelsep=colon,strut=off} % DO NOT CHANGE THIS
\floatstyle{ruled}
\newfloat{listing}{tb}{lst}{}
\floatname{listing}{Listing}
\title{Sweeping Heterogeneity with Smart MoPs: \\ Mixture of Prompts for LLM Task Adaptation}
\author {
    Chen Dun\textsuperscript{\rm 1},
    Mirian Hipolito Garcia\textsuperscript{\rm 2},
    Guoqing Zheng\textsuperscript{\rm 2},
    Ahmed Hassan Awadallah\textsuperscript{\rm 2},
    Anastasios Kyrillidis\textsuperscript{\rm 1},
    Robert Sim\textsuperscript{\rm 2}
}
\affiliations {
    \textsuperscript{\rm 1}Rice University\\
    \textsuperscript{\rm 2}Microsoft\\
}

\usepackage{bibentry}
\begin{document}

\maketitle

\begin{abstract}
Prompt instruction tuning is a popular approach to better adjust pretrained LLMs for specific downstream tasks. 
How to extend this approach to simultaneously handle multiple tasks and data distributions is an interesting question. 
We propose \emph{Mixture of Prompts} (MoPs) with smart gating functionality. 
Our proposed system identifies relevant skills embedded in different groups of prompts and dynamically weighs experts (i.e., collection of prompts) based on the target task. 
Experiments show that MoPs are resilient to model compression, data source, and task composition, making them highly versatile and applicable in various contexts. 
In practice, MoPs can simultaneously mitigate prompt training ``interference'' in multi-task, multi-source scenarios (e.g., task and data heterogeneity across sources) and possible implications from model approximations. 
Empirically, MoPs can reduce final perplexity from 9\% up to 70\% in non-i.i.d. distributed cases and from 3\% up to 30\% in centralized cases, compared to baselines.
\end{abstract}


\section{Introduction}
\label{sec:intro}

%\textbf{Background.} 
Advances in large language models (LLMs) show that they are powerful general-purpose models \citep{brown2020language,bommasani2021opportunities, bubeck2023sparks},
with the ability to solve different tasks out of the box. 
(Soft) prompt instruction tuning helps in this direction by finetuning deployed models to adjust to individual downstream tasks, without the need of full-model finetuning \citep{lester2021power, NEURIPS2022_b1efde53, kenton2021alignment,10.1145/3442188.3445922, tamkin2021understanding}.\footnote{Similar attempts created the term \textit{parameter-efficient finetuning} (PEFT) methods \citep{houlsby2019parameter, ding2023parameter}, including adapter tuning \citep{houlsby2019parameter, hu2023llm}, prefix tuning \citep{li2021prefix}, prompt tuning \citep{lester2021power}, low-rank adaptation (LoRA) \citep{hu2021lora}, few-shot tuning (IA$^3$) \cite{liu2022fewshot} and compression aware prompts \citep{xu2023compress}; in this work, without loss of generality and as a proof-of-concept, we focus on soft prompt instruction pruning. } %among others.

%\textbf{A gap that persists.} 
Yet, vanilla finetuning procedures do not always behave well in %the applications in mind in the above studies do not correspond to 
some practical scenarios. 
For instance, consider the following: 
A company $\texttt{XYZ}$ intends to develop a general-purpose office assistant application that solves different tasks at the same time.
A desired solution would be to create a family of prompts that can be combined on-the-fly and tackle a wide variety of incoming tasks, without heavy supervision.
With that goal in mind, company $\texttt{XYZ}$ uses both human ``labelers'' to generate demonstration data of related tasks (maybe, stored in a central server) and client data to utilize their own local demonstration.

Most existing approaches suggest the ``centrally training and finetuning'' solution, where all data (company-owned and private client) are gathered, and a pre-trained model is further finetuned. 
However, beyond the privacy concerns due to centrally collecting all data, such an approach, while it would work for the tasks and data included in the training dataset, it would not necessarily adapt to new incoming tasks. 
Further, company $\texttt{XYZ}$ desires to decrease both training and inference costs of the final model by deploying highly compressed models and aiming for the generation of specialized ``experts'' that can be used on the fly and just-in-time for most incoming clients, without necessarily requiring further finetuning.\footnote{The definition of an ``expert'' here will be apparent later in the text; this should not necessarily be assumed as MLP experts in a sparse mixture of expert scenarios \citep{puigcerver2023sparse}.}

%What company $\texttt{XYZ}$ is facing is the following challenge: 
%\noindent \emph{Can existing prompt-tuning strategies utilize all the available data from central servers and local clients to construct specialized experts --instead of randomized ones-- while maintaining desirable computation/communication costs?} 

\textbf{Overview of our approach and contributions.} 
Such scenarios suggest a \textit{multi-source, multi-task prompt tuning} approach. %, including centralized training and federated learning scenarios. 
We consider scenarios where the system does not just face data from a single data distribution but has to learn how to handle data and task heterogeneity.
Following recent literature \cite{mittal2022modular}, our emphasis here is on training specialized prompts that operate in a modular way such that, when combined, they tackle tasks in a just-in-time manner.

\begin{figure}[ht]
\begin{center}
\centerline{\includegraphics[width=0.8\columnwidth]{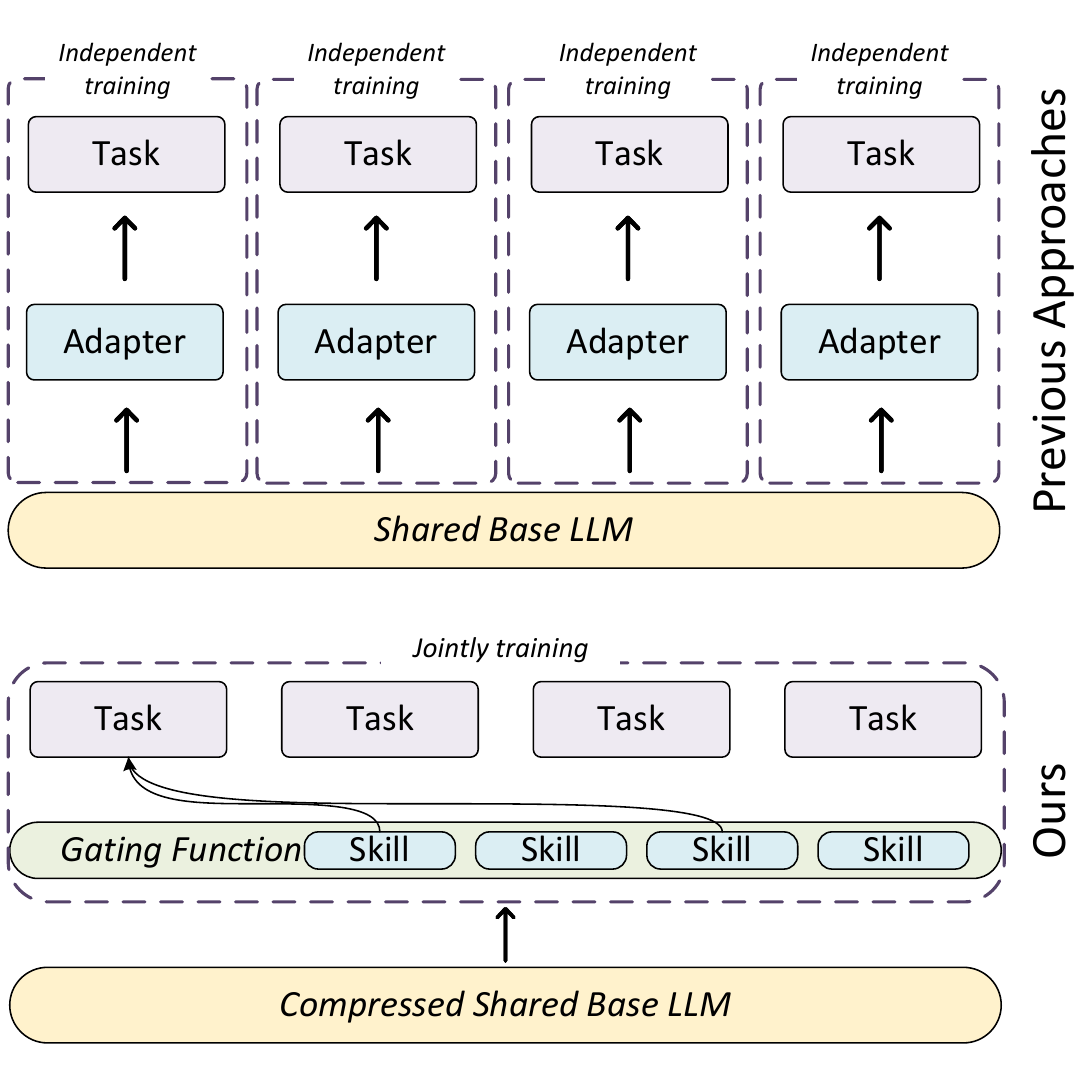}}
\vspace{-1em}
\caption{Multi-Source Multi-task Training}
\label{fig:sce}
\vspace{-2em}
\end{center}
\end{figure}

We propose to use \textit{Mixture of Prompts} (or MoPs) in \emph{multi-source, multi-task prompt instruction tuning} to efficiently leverage all available data from both the central server and local clients while using highly compressed models. See Figure \ref{fig:sce} for an overview of the differences between our system and dominating existing approaches. 

The use of MoPs is guided by a gating functionality that can identify relevant skills embedded in different groups of prompts (``experts'' in this work) based on the data domain of the current input. 
This approach dynamically selects a ``soft'' combination of relevant prompts, which we hypothesize is critical to avoid the appearance of implicit ``interference'' during training. Simply put, if prompts are not aware of multiple data sources and we allow all prompts to ``follow" (=be trained toward) the streaming incoming tasks and data distributions, then all prompts would learn the same things and lead to plain ``average'' solutions during training as data and tasks change.
The latter is similar to ``simple aggregation'' rules using PEFT techniques, where one averages the updated prompts.
In contrast, we propose:
\vspace{-0.0cm}
\begin{itemize}[leftmargin=*]
    \item \textbf{Tackling task/data heterogeneity.} MoPs could utilize either solely centralized data collected by human ``labelers'', heterogeneous local data (e.g., stored on edge devices), or a combination of those while being agnostic about the composition of instruction data. 
    \vspace{-0.1cm}
    \item \textbf{Adaptable Computing.} The design of MoPs enables flexibility in the model architecture, as the prompts (experts) can be placed in any layer rather than just the first layer. This reduces the computational cost of training, as fewer layers need to be backpropagated.
    \vspace{-0.1cm}
    \item \textbf{Model compression resiliency.} We observe an emerging ability of MoPs: they often work \emph{out of the box}, regardless of reasonable model compression ratios or techniques (i.e., pruning, quantization). We note that when the model is not as strong as the full precision model, it is even more difficult for prompts to be trained correctly and learn diverse things. MoPs improve existing baselines, demonstrating their effectiveness and robustness. 
    \vspace{-0.1cm}
    \item \textbf{Empirical performance.} MoPs manage to decrease final perplexity from $\sim9\%$ up to $\sim70\%$, as compared to baselines in the federated scenario, and from $\sim 3\%$ up to $\sim30\%$ in the centralized scenario. Our gains in the non-i.i.d. (federated) setup support our hypothesis that MoPs effectively improve upon data heterogeneity under skewed distributions, thus reducing the model drift problem.  \vspace{-0.1cm}
\end{itemize}

\section{Related Works}
While multi-task learning and multi-source learning in LLMs have been considered in the past \citep{radford2021learning, reed2022generalist, huang2023language, bubeck2023sparks}, to our knowledge, there is limited work on PEFT methods that satisfy the above desiderata. 

From the federated learning perspective, \citep{babakniya2023slora, chen2023federated} considers the federated version of LoRA \citep{hu2021lora}; \citep{zhang2023fedyolo} considers the federated version of adapters;
while \citep{jiang2023fdapt} suggests ongoing pretraining of the full model for better domain adaptation, based on the findings in \citep{gururangan2020don}.
Yet, to our understanding, \textit{these works focus primarily on the periodic aggregation and averaging of the PEFT-based parameters}, without targeting on specialized experts. %  that --when combined-- outperform on just-in-time tasks, based on compressed models. % and working in a modular way. 
Concurrent work on multiple prompts \citep{si2023mixture, asai2022attentional} assumes a prior knowledge of skills/tasks and uses hand-designed ``expert'' prompts.  
These do not consider multi-source data heterogeneity. \vspace{0.1in}%while \citep{si2023mixture} uses a random forest as a gating function. 

Inspired by studies on the linear connectivity of trained models in a full finetuning setting \cite{wortsman2022model, matena2022merging, jin2022dataless, ainsworth2022git}, \cite{zhang2023composing} considers composing PEFT modules via linear arithmetic operations, drawing connections with the word-embedding hypothesis (like the famous ``\texttt{queen = king - man + woman}'' equation \cite{mikolov2013distributed}). 
Such efforts are interesting but orthogonal to this work and could be combined as a future direction; in comparison, our gating function weighs and combines prompts (``experts'') on the fly, where their weights can be added or subtracted in a learnable way, potentially providing more flexibility in how modules are combined.\vspace{0.1in}

\texttt{AdapterSoup} \cite{chronopoulou2023adaptersoup} averages the weights of the adapters that are most related to the new domain to improve out-of-domain
performance at test time.
This resembles FedAvg in federated learning \cite{mcmahan2017communication}, where the weights used to aggregate models (here, PEFT modules) are not necessarily uniform. 
\cite{chronopoulou2023adaptersoup} use text clustering and semantic similarity on language tasks to compute the aggregation weights during testing; in our scenario, we are learning the gating function weights to automatically decide which experts and how much they should be combined.\vspace{0.1in}

\cite{zhang2023increlora} considers the incremental parameter allocation in LoRA to accommodate finetuning in new tasks. 
To improve adapter capacity without increasing parameters or computational cost, \texttt{AdaMix} in \cite{wang2022adamix} introduces multiple shared adapter components, with sparse random routing during training, that are later on merged via averaging to a single PEFT module; there, the authors do not consider the case of multiple-source/multiple-task scenarios, but focus on the efficiency component. 
\texttt{AdaFusion} in \cite{pfeiffer2021adapterfusion} learns to combine adapters specific to different tasks, as well as a shared pretrained model, by introducing an additional attention layer that fuses all the above during additional training. These parameters learn to combine the adapters as a dynamic function of the target task data.
\texttt{SMEAR} in \cite{muqeeth2023soft} uses a given router’s distribution to average the parameters of the corresponding experts and then routes the input through a single merged expert, which is better than discrete routing.\vspace{0.1in}

\color{black}
In \cite{mahabadi2021parameter}, the authors employ adapter modules within the layers of a pretrained model. Table 4 compares our method to a similar technique, such as LoRA, a widely adopted method that involves efficient finetuning by updating the model weights. Our results show that MoPs show merit compared to prior work.

In \cite{wu2023parameter}, the authors create a bank of prompts, which integrates cross-task features from diverse sources, serving as an essential component for initializing the prompts before finetuning specific tasks. In contrast, MoPs learn task similarities from scratch under the supervision of a gating function that learns to scale only the relevant experts according to the current input.

\section{Background}
\label{sec:definition}
\textbf{LLMs and Decoder-only Transformers.} The backbone of LLMs are decoder-only transformers \citep{vaswani2017attention, liu2018generating}. 
An LLM takes as input a question/context and performs next-word prediction to generate responses to the question. 
The forward pass of the $\ell$-th layer is shown below; for all head indices $h \in \{1, 2, \dots, H\}$: 
\begin{align*}
    \mathbf{Q}^{h, \ell} &= \mathbf{W}_q^{h, \ell}\mathbf{X^{\ell}} \in \mathbb{R}^{d_h \times n} \\
    \mathbf{K}^{h, \ell} &= \mathbf{W}_k^{h, \ell}\mathbf{X^{\ell}} \in \mathbb{R}^{d_h \times n} \\
    \mathbf{A}^{h, \ell} &=\texttt{Softmax}\left(\mathbf{M} \odot \left(\mathbf{Q}^{h, \ell \top} \mathbf{K}^{h, \ell}\right)\right) \in \mathbb{R}^{n \times n}; \\
    \mathbf{V}^{h, \ell} &= \left(\mathbf{W}_v^{h, \ell} \cdot \mathbf{X}^{\ell}\right) \cdot \mathbf{A}^{h, \ell}  \in \mathbb{R}^{d_{h} \times n}; \\
    \mathbf{O}^{\ell} &= \mathbf{W}_o^{\ell} \cdot \texttt{Concat}\left(\mathbf{V}^{1, \ell}, \dots, ~\mathbf{V}^{H, \ell} \right) \in \mathbb{R}^{d_{o}\times n}; \\
     \mathbf{X}^{\ell+1} &= \mathbf{W}_{\text{ff2}}^\ell(\texttt{Relu}(\mathbf{W}_{\text{ff1}}^\ell\mathbf{O}^\ell)) \in \mathbb{R}^{d_t \times n}.
\end{align*} 
In particular, let $d_{h}$ be the dimension of the attention head, $d_{t}$ the dimension of the input token embedding, $d$ the width of the feedforward layer, $H$ the number of attention heads, and $n$ the input sequence length. 
Here, $\texttt{Concat}(\mathbf{B}, \mathbf{C})$ --with $\mathbf{B}$ and $\mathbf{C}$ of appropriate dimensions-- concatenates the two matrices columnwise.
In the above expressions, $\mathbf{X^{\ell}} \in \mathbb{R}^{d_t \times n}$ is the input to the $\ell$-th layer (i.e., when $\ell = 0$, this is the data input); $\mathbf{W}_{q}^{h, \ell}, \mathbf{W}_{k}^{h, \ell}, \mathbf{W}_{v}^{h, \ell} \in \mathbb{R}^{d_h \times d_t}$ are the weight matrices associated with the query, key and value embedding inputs, where we use for simplicity the same dimension $d_h$ for all of them; $\mathbf{W}_{o}^{\ell} \in \mathbb{R}^{d_o \times (H\cdot d_h)}$ is the weight matrix associated with the output of the attention head before the FFN layer, $\mathbf{M} \in \mathbb{R}^{n \times n}$ is the decoder attention mask that prevents positions from attending to the future.
Finally, $\mathbf{W}_{\text{ff1}}^\ell \in \mathbb{R}^{d_1 \times d_o}$ and $\mathbf{W}_{\text{ff2}}^\ell \in \mathbb{R}^{d_t \times d_1}$ are the weight matrices of the fully-connected layers.\vspace{0.1in}

\textbf{LLMs with Trainable Prompts:}
Following \citep{NEURIPS2022_b1efde53, kenton2021alignment,10.1145/3442188.3445922, tamkin2021understanding}, we consider trainable (soft) prompts to perform efficient instruction tuning on LLMs. 
Using similar notation and additional $K$ trainable prompts \textcolor{blue}{$\mathbf{P}^\ell \in \mathbb{R}^{d_{t} \times K}$}, for some $\ell \in \{1, \dots, L\}$, the forward pass of the $\ell$-th module where prompts apply can be formulated as below (\textcolor{blue}{blue} text highlights the main differences with the above): \vspace{-0.6cm}
{\small
\begin{flalign*}
    &\textcolor{blue}{\mathbf{B}^{\ell} = \texttt{Concat}(\mathbf{P}^{\ell},\mathbf{X}^{\ell}) \in \mathbb{R}^{ d_t \times (n+K)}} & \\
    % &\mathbf{C}^{\ell} = \texttt{Concat}(\mathbf{P}^{\ell},\mathbf{X}^{\ell}) \in \mathbb{R}^{ d_t \times (n+K)} & \\
    &\mathbf{Q}^{h, \ell} = \mathbf{W}_q^{h, \ell} \textcolor{blue}{\mathbf{B^{\ell}}} \in \mathbb{R}^{d_h \times (n \textcolor{blue}{+ K})} & \\
    &\mathbf{K}^{h, \ell} = \mathbf{W}_k^{h, \ell} \textcolor{blue}{\mathbf{B^{\ell}}} \in \mathbb{R}^{d_h \times (n \textcolor{blue}{+ K})} & \\
    &\mathbf{A}^{h, \ell} = \texttt{Softmax}\left(\mathbf{M}' \odot\left(\mathbf{Q}^{h, \ell \top}\mathbf{K}^{h, \ell}\right)\right) \in \mathbb{R}^{(n\textcolor{blue}{+K}) \times (n\textcolor{blue}{+K})}; & \\
    &\mathbf{V}^{h, \ell} = \left(\mathbf{W}_v^{h, \ell} \cdot \textcolor{blue}{\mathbf{B}^{\ell}}\right) \cdot \mathbf{A}^{h, \ell}  \in \mathbb{R}^{d_{h} \times (n \textcolor{blue}{+ K})}; & \\    
    &\mathbf{O}^{\ell} = \mathbf{W}_o^{\ell} \cdot \texttt{Concat}\left(\mathbf{V}^{1, \ell}, \dots, ~\mathbf{V}^{H, \ell} \right) \in \mathbb{R}^{d_{o}\times (n \textcolor{blue}{+ K})}; & \\
    &\textcolor{blue}{\texttt{Concat}(\mathbf{P}^{\ell+1},\mathbf{X}^{\ell+1})} = \mathbf{W}_{\text{ff2}}^\ell(\texttt{Relu}(\mathbf{W}_{\text{ff1}}^\ell\mathbf{O}^{\ell})) \in \mathbb{R}^{d_t \times (n + K)}, &
\end{flalign*}
}
%\vspace{-0.2cm}
where $\mathbf{W}^{h, \ell}_q$, $\mathbf{W}^{h, \ell}_k$, $\mathbf{W}^{h, \ell}_v$, $\mathbf{W}_{\text{ff1}}^\ell$, $\mathbf{W}_{\text{ff2}}^\ell$ are all \textit{frozen}.
How $\mathbf{P}^\ell$ are treated (i.e., frozen or trainable) for different $\ell$ values will be described in the text.
%After the first layer, we treat $\mathbf{P}^{\ell}$ as regular tokens embedding through LLM layers. 
$\mathbf{M'} \in \mathbb{R}^{(n + K) \times (n + K)} $ is the modified decoder attention mask where all prompts are never masked out for all input tokens. 
We omit skip connections and layer normalization to simplify notations. 

\textbf{Injection of prompts.} 
Inspired by \citep{li2021prefix, liu2022late}, we propose next to reduce communication costs by injecting trainable prompts in the intermediate layers. By enabling the insertion of these prompts on different layers, we can significantly reduce the backpropagation time, thus improving the speed and efficiency of the training process. As illustrated in Figure \ref{fig:overview}, this approach allows for an adjustable design tailored to different setups.

\textbf{Prompt-tuning in Federated Learning:}
Recent approaches adapt FedAvg \citep{mcmahan2017communication} to prompts tuning \citep{zhao2023fedprompt,babakniya2023slora}: 
%During the local training phase, each client will optimize the local copy of prompts. 
During synchronization, all updated copies of prompts are averaged on the server for the next round of training. 
\textit{This is in contrast with this work: while the idea of mixing prompts is not new,  we are focusing on learning relevant skills as expressed via selected subsets of prompts based on the data domain of the current input and dynamically selecting the combination of appropriate prompts to solve current and new tasks.} 

\section{MoPs with a smart gating function} 
\label{sec:fed_moe}

\textbf{Our hypotheses in a nutshell.} 
Training prompts to handle \emph{universally} multi-source/multi-task scenarios might result in \textit{prompt interference} across tasks and sources. 
%Current prompt tuning approaches (both centralized and federated) might not operate to their full potential, especially when facing task heterogeneity (i.e., when training involves multiple tasks simultaneously), data heterogeneity (i.e., when training with imbalanced data, e.g., across distributed clients), and when approximate (e.g., compressed) models are in use to further reduce computation costs. 
As our hypothesis, a way \textit{prompt interference} can be decomposed is: 
%\vspace{-0.3cm}
\begin{itemize}[leftmargin=*]
    \item In centralized training, prompts might converge to poor-performing parameter configurations when heterogeneous tasks are considered due to \emph{conflicting training signals from different tasks} \cite{vzliobaite2010learning}. This case is challenging when the tasks are distinctly diverse. \vspace{-0.05cm}
    \item The above resembles the scenario of \textit{model drifting} in federated learning \cite{li2020federated, karimireddy2020scaffold, li2021model}: when different clients ``pull'' the trainable model to work well on their local data, the aggregated model across clients results in a poor-performing final model. In such privacy-preserving scenarios, \emph{heterogeneous data distributions} add more training interference across clients. The model can be biased towards the tasks with more data, losing its capability for generalization. \vspace{-0.05cm}
    \item  For efficiency reasons, compressed LLMs \cite{frantar-sparsegpt, frantar2022gptq, sun2023simple, jaiswal2023emergence, ji2023pruning, ma2023llm, liu2023llm, kim2023memory, kim2021bert, dettmers2023spqr, dettmers2023qlora, lin2023awq} are now widely used for both centralized and FL scenarios. Such model approximations could impose implicit prompt training interference (and thus leave room to improve performance), as trainable prompts are responsible for \emph{both recovering model capacity loss and model adaptation for downstream tasks}.  \vspace{-0.05cm}
\end{itemize}
%\vspace{-0.8cm}

% \textbf{Algorithm desiderata.} The designed methodology should: 
% %Thus, our key insights are as follows: (1) 
% $i)$ be able to learn from scratch a diverse set of ``skills'' that will be embedded in different prompts to avoid interference or help to recover such ``suppressed'' skills due to model compression; 
% $ii)$ dynamically select and combine the prompts with relevant skills for any incoming input data.
% The latter is in contrast to existing methods \citep{babakniya2023slora, chen2023federated, zhang2023fedyolo, jiang2023fdapt} that often use all prompts (or an aggregation of all prompts) for all subtasks during training and testing.

\subsection{Mixture of Experts (MoPs) Design}

\begin{figure}[ht]
\begin{center}
\centerline{\includegraphics[width=1.\columnwidth]{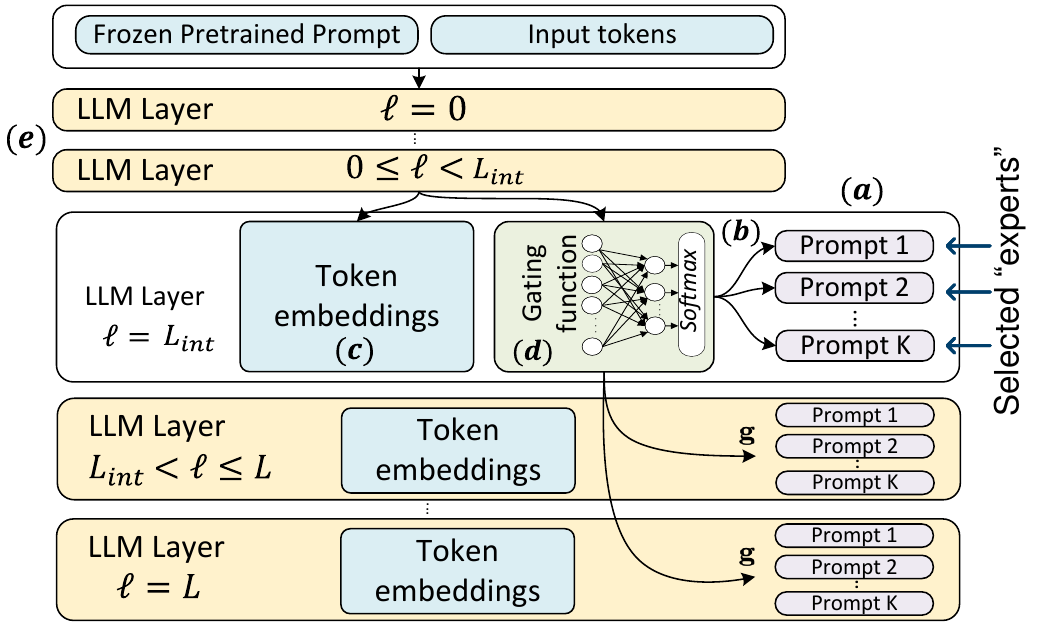}}
\caption{Mixture of prompts with a smart gating function on compressed LLMs overview.}
\label{fig:overview}
\vspace{-0.3cm}
\end{center}
\end{figure}

\textbf{Half through the procedure: Basics \& Pretraining.}
Consider $\mathbf{P}^1 \in \mathbb{R}^{d_t \times K}$ as some frozen pretrained prompts at the input layer, and $\mathbf{X}^1$ is the input data. 
I.e., we assume that either $\mathbf{P}^1$ is provided to us or we first train $\mathbf{P}^1$ for a few epochs using the basic neural network (i.e., without the use of mid-prompt injection or any gating function). 
This means that $\mathbf{B}^{1} = \texttt{Concat}(\mathbf{P}^{1},\mathbf{X}^{1}) \in \mathbb{R}^{ d_t \times (n+K)}$ in the description in Section 2 represents the concatenation of the input tokens with these frozen pretrained prompts; see Figure \ref{fig:overview}(e).
Note also that $\mathbf{W}^{h, \ell}_q$, $\mathbf{W}^{h, \ell}_k$, $\mathbf{W}^{h, \ell}_v$, $\mathbf{W}_{\text{ff1}}^\ell$, $\mathbf{W}_{\text{ff2}}^\ell$ are all \textit{pretrained and frozen}.

Each layer $\ell$ of our architecture follows the description in the \textbf{LLMs with Trainable Prompts} Section, leading to the output $\texttt{Concat}(\mathbf{P}^{\ell+1},\mathbf{X}^{\ell+1})$, where the latter is split into ``prompts'' and regular output, based on their positions. 
We consider the representation $\mathbf{P}^{\ell}, \ell > 1$ as the ``transformed prompt'': i.e., the result as we propagate prompts through the layers. \textit{We use this split for ease of notation and consistency among layers}.
Note that, while $\mathbf{P}^1$ is trainable during the pretraining phase, the rest $\mathbf{P}^{\ell}, ~\ell > 1$ represent the output of intermediate layers, and are not trainable. 
See also the first part of Algorithm \ref{alg:gate}.

\textbf{During MoP training: Prompts as experts.} 
Given this initial setting where we have $\mathbf{W}^{h, \ell}_q$, $\mathbf{W}^{h, \ell}_k$, $\mathbf{W}^{h, \ell}_v$, $\mathbf{W}_{\text{ff1}}^\ell$, $\mathbf{W}_{\text{ff2}}^\ell$ and $\mathbf{P}^{1}$ \textit{pretrained and frozen}, we apply \textit{prompt injection} \citep{li2021prefix, liu2022late}: at an intermediate layer level $L_{\text{int}} \in \{1, \dots, L\}$, we inject multiple (say, $K$) trainable prompts as experts to embed different skills across subtasks, each specializing in different skills; see Figure \ref{fig:overview}(a).
These prompts are trainable in the following sense: 
At this training phase, they are weighted by a gating function (see Figure \ref{fig:overview}(b) and (d), and as described below), depending on the current input. 
The input to the gating function is designed to be the embedding of the network's input via the first $\ell < L_{\text{int}}$ layers; see Figure \ref{fig:overview}(c).\footnote{Beyond a design choice, we inject the prompts in the intermediate layers as a means to reduce the training cost. An alternative --and fully valid-- procedure would be to use two LLMs, where the first is used as an embedding mechanism, and the other uses the embeddings as inputs to the gating function. To perform fair comparisons in our experiments, we decided not to change the size of a given LLM but rather change its design in this way.}
Prompts are selected and updated based on the output weights of the gating function; these weights mitigate the training interference between prompts. 
This allows us to use different combinations of skills for various input tasks. %, resulting in a more accurate handling of incoming tasks. 

\textbf{The gating function.} 
To dynamically weigh prompts based on the input, we have designed a gating function that embeds the present question. 
When $\ell = L_{\text{int}}$, the embedding $\texttt{Concat}(\mathbf{P}^{\ell+1},\mathbf{X}^{\ell+1}) \in \mathbb{R}^{d_t \times (n + K)}$ gets averaged wrt its ``sequence length'' to obtain $\texttt{avg\_emb} \in \mathbb{R}^{d_t}$; see Algorithm \ref{alg:gate}.
I.e., to avoid incurring extra computation and memory costs, as is common in MoEs, our gating function directly utilizes the first part of the given model ($1 \leq \ell \leq L_{\text{int}}$) as the embedding network, without any additional cost. 

The gating function comprises a shallow MLP network that takes the intermediate layer (averaged) embeddings as input, followed by a softmax layer that outputs expert scores: \vspace{-0.2cm}
\begin{align*}
    \mathbf{g} &= \texttt{Softmax}(\mathbf{W}_{\text{gff2}}(\texttt{Relu}(\mathbf{W}_{\text{gff1}}(\texttt{avg\_emb})))) \in \mathbb{R}^K. 
\end{align*}
$\mathbf{W}_{\text{gff1}} \in \mathbb{R}^{d_g \times d_t}$ and $\mathbf{W}_{\text{gff2}} \in \mathbb{R}^{K \times d_g}$ are trainable weights. 

\textbf{After the intermediate layer.} 
Scores in $\mathbf{g} \in \mathbb{R}^K$ are used to scale the network's attention over the collection of experts in the subsequent layers ($L_{int}\leq \ell \leq L$), as shown in Figure \ref{fig:overview}(d). 
In particular, these scores scale accordingly the weights of the prompt embeddings $\mathbf{B}^{\ell} = \texttt{Concat}(\widehat{\mathbf{P}}^{\ell}, \mathbf{X}^{\ell})$, so that those related to the task at hand are more crucial than the others, shown mathematically as follows\footnote{\textcolor{black}{While a vector/matrix can always be represented as a linear combination of fixed vectors, it is often the case that such a family of vectors should be the \textit{basis vectors}. I.e., while it seems as an alternative to use specific weights over a set of fixed vectors to result in e.g., $v = \alpha_1 v_1 + \alpha_2 v_2 + \dots$, such an approach would either require many ``basis'' prompts to be combined together and/or more training iterations, as the prompts are fixed and do not provide additional flexibility to achieve the desiderata.}}:
\begin{align*}
    \widehat{\mathbf{B}}^{\ell} = \mathbf{B}^{\ell}_{:, 1:K} \odot \mathbf{g} \in \mathbb{R}^{d_t \times (n + K)}, \quad \forall L_{\text{int}} \leq \ell \leq L. 
\end{align*}
Here, $\odot$ represents the Hadamard product; note that $\mathbf{B}^{\ell}_{:, 1:K}$ represents the $K$ first columns of $\mathbf{B}^{\ell}$. We overload the Hadamard multiplication to indicate that $\mathbf{g}$ applies to all the rows of this restricted matrix, recursively.
This is highlighted in Algorithm \ref{alg:gate}. 
Using softmax output score, the gating function ``forces'' later layers to focus mostly on dominating selected prompts, which scales the updates for each prompt accordingly during backpropagation. Our gating function imposes a negligible computation overhead in total.

\textbf{Pretraining the gating function.}
To improve the initial performance of our gating function, we assume we have unlabeled instruction data (instruction/question only) with domain/task labels on the server side. 
As such data are instruction only, we can collect such data in both centralized and federated learning cases beforehand. %\footnote{We leave for future study in a more strict federated learning scenario where such unlabeled instruction is also federated.}

To use this data, we manually assign a one-to-one relationship between each prompt group and each data domain/task. 
This provides a good initialization to the gating function, assuming that $i)$ each subtask is drastically different and represents one distinct skill, and $ii)$ each prompt embeds the corresponding skill. 
Such an assumption does not need to be accurate for the available dataset. 
As shown in the experiments, such initialization is good enough: the gating function, together with trainable prompts, can discover a more accurate relationship between subtasks. 

\begin{algorithm*}[ht]
\caption{Mixture of Prompts (MoPs) with a smart gating function}
\label{alg:gate}
\vspace{-0.4cm}
\begin{multicols}{2}
\begin{algorithmic}[l]
    \small 
    \STATE {\textbf{Input}: $\mathbf{W}^{h, \ell}_q$, $\mathbf{W}^{h, \ell}_k$, $\mathbf{W}^{h, \ell}_v$, $\mathbf{W}_{\text{ff1}}^\ell$, $\mathbf{W}_{\text{ff2}}^\ell, \ell \in [L]$ pretrained and frozen; $\mathbf{P}^1$ pretrained and frozen (see text); $L_{\text{int}} \in \{1, \dots, L\}$; input data $\mathbf{X}^1$.}
    \STATE {\textbf{Details}: $\odot$ denotes row-wise element; we inject prompts $\widehat{\mathbf{P}}^{L_{\text{int}}}\in \mathbb{R}^{d_t \times K}$ at the layer $\ell = L_{\text{int}}$.}
    \\\hrulefill
    \STATE \quad \quad ~~$\spadesuit$~ \textbf{Before intermediate layer \& gating function}~$\spadesuit$
    \FOR{ $1 \leq \ell < L_{\text{int}}$ }
        \STATE $\mathbf{B}^{\ell} = \texttt{Concat}(\mathbf{P}^{\ell},\mathbf{X}^{\ell})$
        \FOR{ $1 \leq h \leq H$}
                \STATE $\mathbf{Q}^{h, \ell} = \mathbf{W}_q^{h, \ell}\mathbf{B^{\ell}}$
                \STATE $\mathbf{K}^{h, \ell} = \mathbf{W}_k^{h, \ell}\mathbf{B^{\ell}}$
                \STATE $\mathbf{A}^{h, \ell} = \texttt{Softmax}\left(\mathbf{M}' \odot\left(\mathbf{Q}^{h, \ell \top}\mathbf{K}^{h, \ell}\right)\right)$;
        	\STATE $\mathbf{V}^{h,   \ell} = \left(\mathbf{W}_v^{h, \ell} \cdot \mathbf{B}^{\ell}\right) \cdot \mathbf{A}^{h, \ell}$                   
	\ENDFOR
        \STATE $\mathbf{O}^{\ell} = \mathbf{W}_o^{\ell} \cdot \texttt{Concat}\left(\mathbf{V}^{1, \ell}, \dots, ~\mathbf{V}^{H, \ell} \right)$
        \STATE $\texttt{Concat}(\mathbf{P}^{\ell+1},\mathbf{X}^{\ell+1}) = \mathbf{W}_{\text{ff2}}^\ell(\texttt{Relu}(\mathbf{W}_{\text{ff1}}^\ell\mathbf{O}^{\ell}))$  
    \ENDFOR
    \\\hrulefill
    \STATE \quad \quad \quad ~~$\spadesuit$~ \textbf{Intermediate layer \& gating function}~$\spadesuit$
    \IF{ $\ell=L_{\text{int}} $}
    \STATE {Let $\texttt{avg\_emb} \in \mathbb{R}^{d_t}$ such that: \vspace{-0.3cm}$$\vspace{-0.4cm} \texttt{avg\_emb}:= \tfrac{1}{n+K} \sum_{i = 1}^{n + K} \texttt{Concat}(\mathbf{P}^{L_{\text{int}}},\mathbf{X}^{L_{\text{int}}})_{:, i}$$. \vspace{-0.1cm}}
    \STATE {Compute $\mathbf{g} \in \mathbb{R}^K$ through the gating function: \vspace{-0.25cm} \begin{align*}
        \mathbf{g} &= \texttt{Softmax}(\mathbf{W}_{\text{gff2}}(\texttt{Relu}(\mathbf{W}_{\text{gff1}}(\texttt{avg\_emb}))))
    \end{align*} }
    \vspace{-0.6cm}
    \ENDIF
    \\\hrulefill
    \STATE \quad \quad \quad \quad \quad \quad ~~$\spadesuit$~ \textbf{After intermediate layer}~$\spadesuit$
    \FOR{ $L_{\text{int}} \leq \ell < L$ }
        \STATE $\mathbf{B}^{\ell} = \texttt{Concat}(\widehat{\mathbf{P}}^{\ell}, \mathbf{X}^{\ell})$
        \STATE $\textcolor{teal}{\widehat{\mathbf{B}}^{\ell} = \mathbf{B}^{\ell}_{:, 1:K} \odot \mathbf{g}}$
        \FOR{ $1 \leq h \leq H$}
                \STATE $\widehat{\mathbf{Q}}^{h, \ell} = \mathbf{W}_q^{h, \ell}\widehat{\mathbf{B}}^{\ell}$
                \STATE $\widehat{\mathbf{K}}^{h, \ell} = \mathbf{W}_k^{h, \ell}\widehat{\mathbf{B}}^{\ell}$
                \STATE $\widehat{\mathbf{A}}^{h, \ell} = \texttt{Softmax}\left(\mathbf{M}' \odot\left(\widehat{\mathbf{Q}}^{h, \ell \top}\widehat{\mathbf{K}}^{h, \ell}\right)\right)$;
        	\STATE $\mathbf{\widehat{V}}^{h,   \ell} = \left(\mathbf{W}_v^{h, \ell} \cdot \widehat{\mathbf{B}}^{\ell}\right) \cdot \widehat{\mathbf{A}}^{h, \ell}$                   
	\ENDFOR
        \STATE $\widehat{\mathbf{O}}^{\ell} = \mathbf{W}_o^{\ell} \cdot \texttt{Concat}\left(\mathbf{\widehat{V}}^{1, \ell}, \dots, ~\mathbf{\widehat{V}}^{H, \ell} \right)$
        \STATE $\texttt{Concat}(\widehat{\mathbf{P}}^{\ell+1},\mathbf{X}^{\ell+1}) = \mathbf{W}_{\text{ff2}}^\ell(\texttt{Relu}(\mathbf{W}_{\text{ff1}}^\ell \widehat{\mathbf{O}}^{\ell}))$  
    
  %       \vspace{-2em}
		% \STATE \begin{align*}
  %           \mathbf{A}^h &= \texttt{Softmax}\left(\mathbf{M'}(\mathbf{W}^h_q(\texttt{Concat}(\mathbf{\hat{P}}^{\ell},\mathbf{X}^{\ell}))^{\top}\right. \\
  %           &\quad \left.(\mathbf{W}^h_k\texttt{Concat}(\mathbf{\hat{P}}^{\ell},\mathbf{X}^{\ell}))\right) \in \mathbb{R}^{(n+K) \times (n+K)};
  %       \end{align*}
		% \STATE $\mathbf{A}^h [:,0:K] = \mathbf{A}^h [:,0:K] \odot \mathbf {G} \in \mathbb{R}^{(n+K) \times (n+K)}$
		% \STATE $\mathbf{\widehat{V}}^h = \mathbf{A}^h \left(\mathbf{W}^h_v \texttt{Concat}(\mathbf{\hat{P}}^{\ell},\mathbf{X}^{\ell})\right) \in \mathbb{R}^{d_{h} \times (n+K)};$
		% \STATE $\mathbf{O} = \mathbf{W}_o\texttt{Concat}\left(\mathbf{\widehat{V}}^0,~\mathbf{\widehat{V}}^1, \dots, ~\mathbf{\widehat{V}}^H \right) \in \mathbb{R}^{d_{t}\times (n+K)};$
		% \STATE $\texttt{Concat}(\mathbf{\hat{P}}^{\ell+1},\mathbf{X}^{\ell+1}) = \mathbf{W}_{\text{ff2}}(\texttt{Relu}(\mathbf{W}_{\text{ff1}}\mathbf{O}))$
	\ENDFOR
\end{algorithmic}
\end{multicols}
\vspace{-0.3cm}
\end{algorithm*}

\textbf{Using compressed LLMs for efficient prompt tuning.} Compressed LLMs \cite{frantar-sparsegpt, frantar2022gptq, sun2023simple} are widely used for downstream instruction tuning due to training efficiency concerns. 
\textcolor{black}{Overall, given the large scale of LLMs, using them out-of-the-box for deployment scenarios is often infeasible due to their resource-intensive requirements}. %Further, the improvements obtained by using compressed models over their uncompressed counterparts are significant, especially in scenarios where computational and memory resources are limited (e.g., multi-task adaptation on edge devices).} Finally, \textcolor{black}{there is a growing body of literature on various techniques for compressing LLMs, including unstructured pruning, structured pruning, distillation of LLMs, and quantization that support this point. These techniques have effectively reduced LLMs' size and have become the norm in model deployment.}

Our system, depicted in Figure \ref{fig:overview}, follows this paradigm by utilizing \emph{compressed LLMs}. We add prompts only to the middle layers, thus allowing flexibility to avoid backpropagation of the entire model during training.

The above are summarized in Algorithm \ref{alg:gate}. 
Briefly, given an input question, MoP first embeds the question using the first $ < L_{\text{int}}$ layers of a given compressed LLM. 
We set $L_{\text{int}}=10$ for a LLama-7B model with $L = 32$. Such choice of $L_{\text{int}}=10$ is to balance two conflicting requirements: $i)$ we want to inject prompts as late as possible to reduce back propagation cost during training; $ii)$ prompts should be injected early to have more capacity in influencing the pretrained LLM network. In the Experiments Section, we provide a detailed analysis of the trade-offs of injecting prompts on different layers. At layer $L_{\text{int}}$, we inject $K$ trainable prompts. 
The gating network uses the previous layer's embedding to generate expert scores for each prompt based on the input question, which is used to re-scale attention weight from other tokens to those prompts. It follows the normal LLM forward propagation after layer $L_{\text{int}}$.

\section{Experiments}
\label{sec:results}

%We evaluate our method's performance and effectiveness on various tasks and contrast it with different baseline approaches. 
%Approximate LLMs will become increasingly valuable due to faster training and inference times and the significant reduction in energy consumption; our results correspond to different pruning ratios. 
%We conduct an ablation study over the different components of our method. 

\textbf{Datasets.} We used two datasets: Databricks Dolly 15k \citep{DatabricksBlog2023DollyV2} and Super-Natural Instructions \citep{naturalinstructions}; see Table \ref{tab:dataset}. 
These datasets challenge our method: MoPs must learn and select relevant skills from scratch without prior knowledge of the complex relationships between the subtasks. 
We split the original 5k samples from each dataset into 90\% training and 10\% testing sets. 
We used a batch size of 1 for both training and testing. 
In the federated scenario, we simulated an uneven distribution of data across 100 clients, resulting in different proportions and sizes of data. The batch size remained at 1. 
The distribution of data skew across clients is explained in Appendix A.

\begin{table}[t]
\vspace{-0.0cm}
\caption{Task categories used per dataset}
\label{tab:dataset}
\centering
\small
\resizebox{1\columnwidth}{!}{
\begin{tabular}{p{1cm}p{2cm}p{5.5cm}}
    \toprule
    Dataset & Dolly-15K Instructions & Super-Natural Instructions\\
    \midrule\midrule
    & creative writing & quoref-question-generation\\
    & closed QA & drop-question-generation\\
    & open QA & essential-terms-identifying-essential-words\\
    Subtasks & summarization & add-integer-to-list\\
    & information extraction & evaluation-semantic-relation-classification\\
    & classification & ljspeech-textmodification\\
    & brainstorming & mmmlu-answer-generation-global-facts\\
    \midrule
    Total & 5000 samples & 5000 samples \\
    \bottomrule
\end{tabular}}
\vspace{-0.4cm}
\end{table}

%%%%%%%%%%%%%%%%%%%%%%%%%%%%%%%%%%
\begin{table*}[!ht]
\vspace{-0.2cm}
\caption{Summary of final perplexities $(\downarrow)$ for the centralized scenario on Dolly-15 and Super-Natural datasets.}
\label{tab:cl_results}
\centering
\resizebox{1.7\columnwidth}{!}{
    \begin{tabular}{c| c|c|c|c|c|c|c|c}
    \toprule
    \multicolumn{2}{c|}{ } &\multicolumn{3}{c|}{Unstructured pruning (ratio)} & \multicolumn{4}{c}{Structured pruning (Type \& Ratio)} \\
    \midrule
    Dataset & Method & 90\% & 85\% & 75\%  & 7:8 (87.5\%) & 3:4 (75\%) & 2:4 (50\%) & 4:8 (50\%)\\
    \midrule\midrule
    & Baseline & 52.65 & 18.16 & 8.25 & 70.14 & 9.06 & 3.67 & 3.59 \\
    Dolly-15K & MoPs & 40.34 & 15.04 & 7.24 & 54.97 & 8.08 & 3.54 & 3.59  \\
    \multicolumn{1}{c|}{} & \multicolumn{1}{c|}{Gain $\pm$} & \multicolumn{1}{c|}{\textbf{\textcolor{teal}{+12.31 (30\%)}}} &\multicolumn{1}{c|}{\textbf{\textcolor{teal}{+3.12 (20\%)}}} &\multicolumn{1}{c|}{\textbf{\textcolor{teal}{+1.01 (13\%)}}} & \multicolumn{1}{c|}{\textbf{\textcolor{teal}{+15.17 (27\%)}}} &\multicolumn{1}{c|}{\textbf{\textcolor{teal}{+0.98 (12\%)}}}  &\multicolumn{1}{c|}{\textbf{\textcolor{teal}{+0.13 (4\%)}}} & \multicolumn{1}{c}{\textbf{\textcolor{teal}{+0.17 (5\%)}}} \\
    \midrule\midrule
    & Baseline & 58.47 & 16.50 & 8.54 &  67.86 & 10.64 & 6.01 & 5.90 \\
    Super-Natural & MoPs & 52.86 & 14.59 & 7.80 & 59.80  & 10.05 & 5.79 & 5.73  \\
    \multicolumn{1}{c|}{} & \multicolumn{1}{c|}{Gain $\pm$} & \multicolumn{1}{c|}{\textbf{\textcolor{teal}{+5.61 (11\%)}}} &\multicolumn{1}{c|}{\textbf{\textcolor{teal}{+1.91 (13\%)}}} &\multicolumn{1}{c|}{\textbf{\textcolor{teal}{+0.74 (9\%)}}} & \multicolumn{1}{c|}{\textbf{\textcolor{teal}{+8.06 (13\%)}}} &\multicolumn{1}{c|}{\textbf{\textcolor{teal}{+0.59 (6\%)}}}  &\multicolumn{1}{c|}{\textbf{\textcolor{teal}{+0.22 (4\%)}}} & \multicolumn{1}{c}{\textbf{\textcolor{teal}{+0.17 (3\%)}}} \\
    \bottomrule
    \end{tabular}
}
\label{tab:cl_results_}
\end{table*}
%%%%%%%%%%%%%%%%%%%%%%%%%%%%%

%%%%%%%%%%%%%%%%%%%%%%%%%%%%%%%%%%%%%%%%
\begin{table*}[!ht]
\vspace{-0.2cm}
\caption{Summary of final perplexities $(\downarrow)$ for the federated scenario, using a pool of 100 available clients.}
\label{tab:fl_results}
\centering
\resizebox{1.7\columnwidth}{!}{
    \begin{tabular}{c| c|c|c|c|c|c|c|c}
    \toprule
    \multicolumn{2}{c|}{ } &\multicolumn{3}{c|}{Unstructured pruning (Ratio)} & \multicolumn{4}{c}{Structured pruning (Type \& Ratio)} \\
    \midrule
    Dataset & Method & 90\% & 85\% & 75\%  & 7:8 (87.5\%) & 3:4 (75\%) & 2:4 (50\%) & 4:8 (50\%)\\
    \midrule\midrule
    & FedPrompt & 98.13 & 28.28 & 11.99 & 143.02 & 17.20 & 5.09 & 4.91 \\
    Dolly-15K & MoPs & 65.25 & 20.77 & 9.45 & 84.10 & 12.20 & 4.23 & 4.06  \\
    \multicolumn{1}{c|}{} & \multicolumn{1}{c|}{Gain $\pm$} & \multicolumn{1}{c|}{\textbf{\textcolor{teal}{+32.88 (50\%)}}} &\multicolumn{1}{c|}{\textbf{\textcolor{teal}{+7.51 (36\%)}}} &\multicolumn{1}{c|}{\textbf{\textcolor{teal}{+2.54 (27\%)}}} & \multicolumn{1}{c|}{\textbf{\textcolor{teal}{+58.92 (70\%)}}} &\multicolumn{1}{c|}{\textbf{\textcolor{teal}{+5.00 (41\%)}}}  &\multicolumn{1}{c|}{\textbf{\textcolor{teal}{+0.86 (20\%)}}} & \multicolumn{1}{c}{\textbf{\textcolor{teal}{+0.85 (21\%)}}} \\
    \midrule\midrule
    & FedPrompt & 76.17 & 18.64 & 9.14 &  91.64 & 14.42 & 6.43 & 6.14 \\
    Super-Natural & MoPs & 66.51 & 16.52 & 7.88 & 72.04  & 12.38 & 5.75 & 5.65  \\
    \multicolumn{1}{c|}{} & \multicolumn{1}{c|}{Gain $\pm$} & \multicolumn{1}{c|}{\textbf{\textcolor{teal}{+9.66 (15\%)}}} &\multicolumn{1}{c|}{\textbf{\textcolor{teal}{+2.12 (13\%)}}} &\multicolumn{1}{c|}{\textbf{\textcolor{teal}{+1.26 (16\%)}}} & \multicolumn{1}{c|}{\textbf{\textcolor{teal}{+19.6 (27\%)}}} &\multicolumn{1}{c|}{\textbf{\textcolor{teal}{+2.04 (16\%)}}}  &\multicolumn{1}{c|}{\textbf{\textcolor{teal}{+0.68 (12\%)}}} & \multicolumn{1}{c}{\textbf{\textcolor{teal}{+0.49 (9\%)}}} \\
    \bottomrule
    \end{tabular}
}\vspace{-0.4cm}
\end{table*}
%%%%%%%%%%%%%%%%%%%%%%%%%%%%%%%%%%%%%%%%

%\vspace{-0.4cm}

\textbf{System.}
We used 4 NVIDIA RTX A6000 GPU with 46 GB Memory. The total training of the prompts took around 2.5 hours (when experts being injected on layer 10th). The federated training was performed in a distributed fashion using 1:1 relationship between expert and worker.

\textbf{Setup.}
We use SparseGPT \citep{frantar-sparsegpt} to perform structured/unstructured pruning and LLM.int8() \cite{dettmers2022llmint8} to perform \texttt{Int8} quantization of the LLama-7B model, creating different compression ratios of the LLM. Inspired by \citep{xu2023compress}, we assign ten prompts to each single expert, totaling seven experts and 70 randomly initialized prompts, ensuring a 1:1 relationship between experts and tasks. Our experiments show that the 1:1 relationship between experts and tasks is not a strict constraint; the gating function adapts to task similarities by dynamically grouping tasks and allocating experts, often using fewer experts than initially assigned (see Appendices B and C for further discussion). 
%This flexibility allows a fixed number of experts to be effective even when the number of tasks is uncertain. 
We also study the impact of varying the number of prompts per expert.

To show our method can further recover/improve the performance of the pruned model, we add such pretrained prompts to both our baselines and our model. We trained these prompts from scratch in a preprocessing step over 20 training steps. These prompts are frozen during training.

In the centralized setting, we use 20000 steps with a learning rate 0.001.
In the federated setting, we adapt FedAvg to average the updated prompts from all active clients during each synchronization round. We use 100 clients, with ten active clients per training round, and set each local training round to 250 training steps. Counting all clients, the total number of training steps is 50000 with a learning rate 0.001. %(Each active client has around 5000 steps in total and ten active clients at each time)  

\textbf{Baselines.} A reasonable baseline is directly applying prompt tuning to centralized and federated training without any gating function. In centralized training, we use the method from \cite{xu2023compress}. 
In federated training, we utilize \texttt{FedPrompt} from \citep{zhao2023fedprompt}, which adapts FedAvg to prompt training and periodically averaging the updated prompts from all clients.
In both cases, to match computation and memory cost with our method during training, we add additional prompts in $L_{\text{int}}=10$ and freeze the given pretrained prompts in the first layer, thus eliminating the need to calculate gradients before $L_{\text{int}}$. 

\textbf{Centralized training results.} 
See Table \ref{tab:cl_results}. $X\%$ denotes the pruned model with $X\%$ weight pruned out for unstructured pruning. For structured pruning, we follow \citep{frantar-sparsegpt} to use $N:M$ to denote pruning $N$ elements out of consecutive $M$ elements in the weight matrix.  
\textcolor{black}{The uncompressed models are already trained extensively on massive data (and actually on data very similar to the scenarios we create for different tasks and data source distributions); thus, one expects these models to work well.}
%Using compressed models is a natural choice that will enable margins of improvement but also corresponds to realistic scenarios, as these models eventually get used in practice. Yet, in 
Table \ref{tab:uncompressed_results} shows gains over other PEFT methods for scenarios where little space exists for improvements.

Our method reduces the final PPL for all cases, with an advantage for the highest pruning ratios. As the pruning ratio increases, more pressure is placed on prompts to recover skills lost due to the model loss. However, MoPs can help to reduce this burden by providing more capacity for task adaptation, thus alleviating the training interference.

\vspace{-0.0cm}
The PPL reduction in the centralized case is more pronounced for unstructured pruning due to lower sparsity. 
\textcolor{black}{Even if one could argue that the model improvements on higher levels might be less useful, in both Tables 2 and 3 we improve the quality of the models by 5 to 10 PPL units in scenarios that matter. Such an improvement is not trivial: as we do not investigate all possible design choices (best optimizer, best tuning, best transformer architecture, etc), we provide one component that can turn a 17PPL loss to 12PPL loss.}

Further analysis of the gating function stages is explored in Appendix B, revealing that during training, the gating function has learned to adjust the distribution of the prompt weight to better specialize the expert assignment.

\begin{figure}[ht]
\vspace{-0.2em}
\begin{center}
\centerline{\includegraphics[width=\columnwidth]{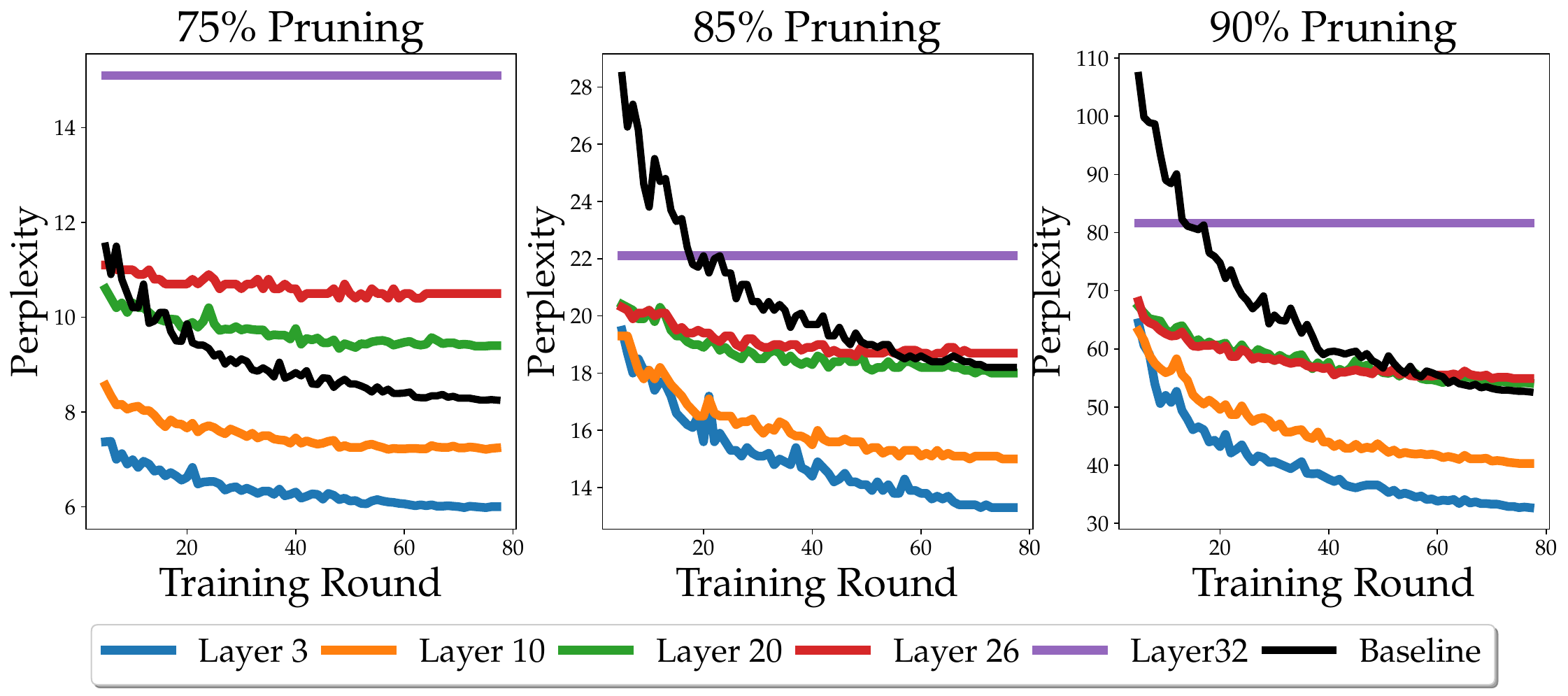}}
\vspace{-0.8em}
\caption{Layer Injection impact on Llama-7B for different unstructured pruning ratios (Dolly-15k) in the centralized setup. Injection on
$L_{\text{int}}=10$ outperforms the baseline. }
\label{fig:diff_layer}
\vspace{-2em}
\end{center}
\end{figure}

\begin{figure}[ht]
\vspace{-0.7em}
\begin{center}
\centerline{\includegraphics[width=\columnwidth]{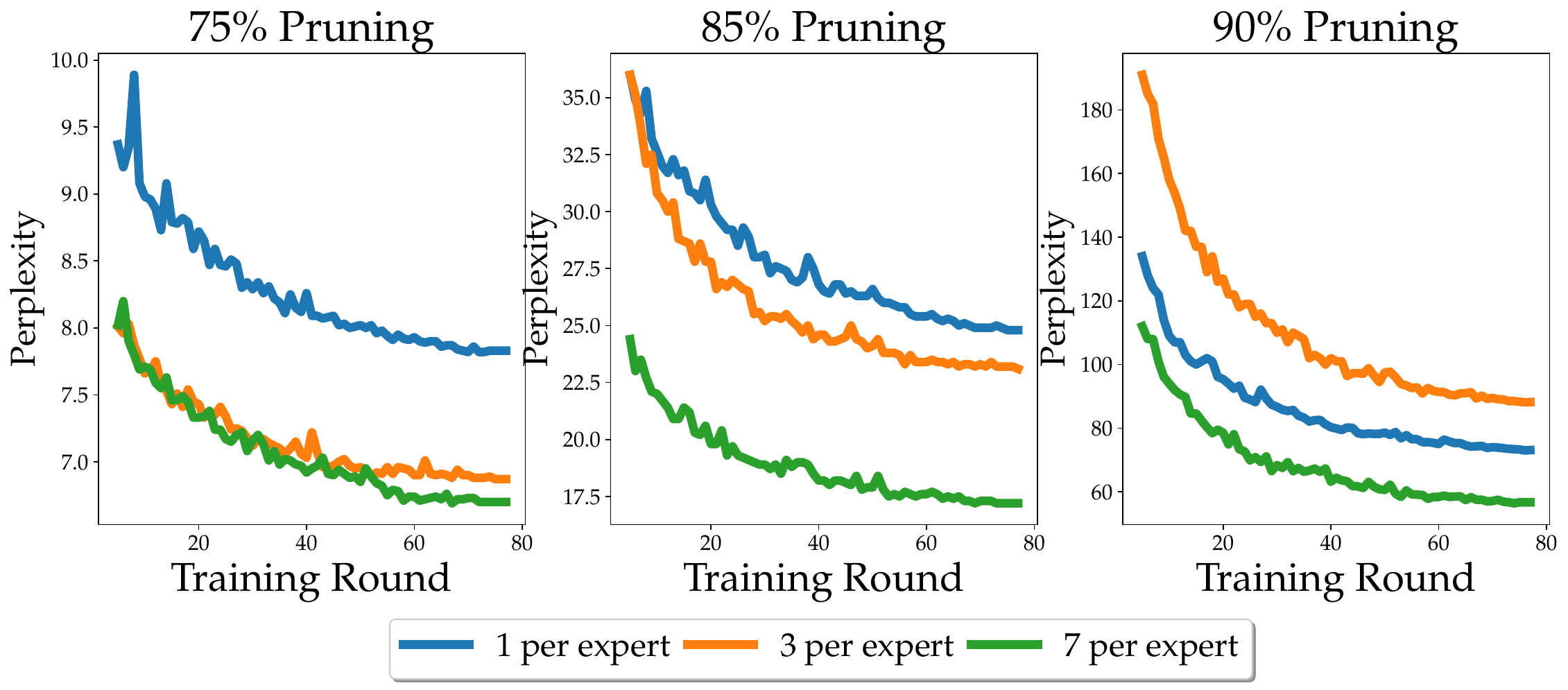}}
\vspace{-0.9em}
\caption{Prompt Injection impact on Llama-7B for different unstructured pruning ratios (Dolly-15k - centralized setup).}
\label{fig:diff_prompts}
\vspace{-3em}
\end{center}
\end{figure}

\textbf{Influence of Injection Layer.} MoP can select the layer where the prompts are injected. We further explored this "hyperparameter" by injecting the experts on different layers. Figure \ref{fig:diff_layer} revealed that injecting the prompts in earlier layers improves the performance, albeit at the cost of ``heavier'' backpropagation. 
%From a cost perspective, it is better to inject them later. However, t
The plots also showed that if the prompts are injected too late, there are only a few layers after the gating function to adjust the prompts, resulting in poorer performance. These results demonstrate the versatility of MoPs, allowing us to adapt the injection as necessary.

\textbf{Influence of Number of Prompts Injected (\textit{Per expert}).} We analyzed how the performance is affected by increasing or decreasing the number of prompts per expert. As illustrated in Figure \ref{fig:diff_prompts}, the more prompts injected, the faster convergence is observed. Providing enough granularity through the prompts allows the gating function to learn more quickly and effectively. Thus, providing sufficient prompts to the experts to maximize the overall performance is essential.

\begin{figure}[ht]
\vspace{-0.6em}
\begin{center}
\centerline{\includegraphics[width=1\columnwidth]{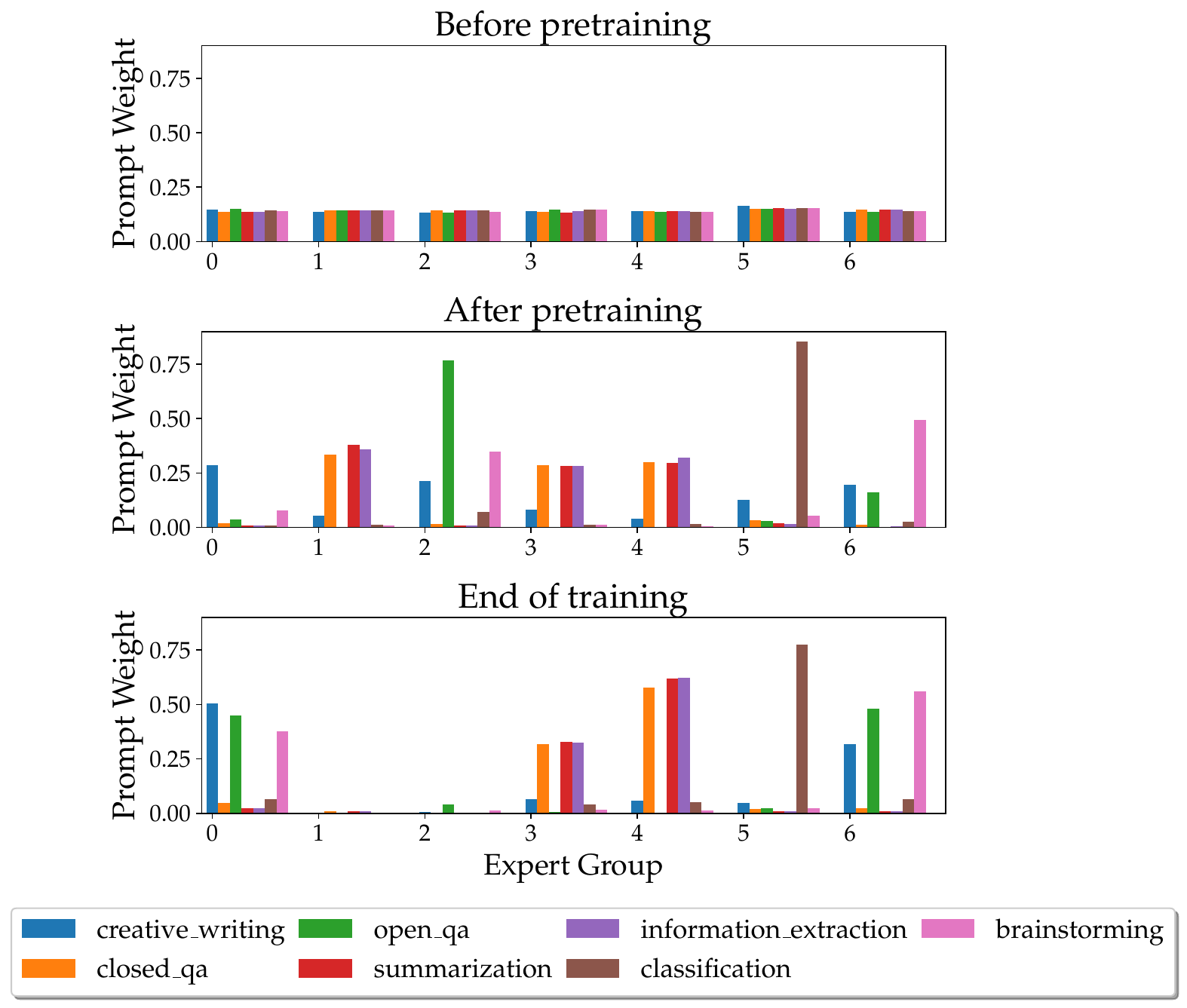}} \vspace{-0.3cm}
\caption{Averaged prompt weights for test dataset using 3:4 (75\%) structured pruning Llama-7B. This verifies how MoPs learn experts with a specialized skill set.}
\label{fig:fl_gate_analysis_75_3_4}
\vspace{-0.8cm}
\end{center}
\end{figure}

\textbf{Federated training results.} Table \ref{tab:fl_results} shows that MoPs is superior to the baseline (\texttt{FedPrompt}) for all pruning ratios. Comparing the relative gains (PPL decrease) with Table \ref{tab:cl_results}, our method yields superior gains in the federated setup. But, \textit{why is MoP performing even better in FL settings?} Figure %\ref{fig:fl_gate_analysis_75_3_4} reveals three distinct phases of the gating function. 
%During the pretraining stage, the gating function can gather useful information about the different tasks, allowing it to capture the domain/task relationships but still be unable to weigh the experts accurately. During the training phase, the gating function selectively updates the relevant experts related to each client, thus ensuring that the prompt updates are properly aligned and preventing the well-known model drift problem in federated scenarios. 
Table \ref{tab:fl_results} demonstrates the gating function's role in overcoming the data heterogeneity, showing improved performance compared to the baseline. 
Figure \ref{fig:fl_gate_analysis_75_3_4} shows how MoPs accentuate the convergence to a set of experts with relatively distinct skills. 
Starting from experts with relatively more mixed skills (after pretraining), it is interesting to note that, at the end of the training, Expert Groups 0 and 6 are specialized in creative writing, open QA, and brainstorming, which all are related to free text generation tasks; Expert Groups 3 and 4 are specialized on closed QA, summarization and information extraction, which all are related to more restricted contexts; finally, the Expert Group 5 is specialized on classification, which is a category by itself.
\textit{This validates our hypothesis and verifies how MoPs learn experts with a specialized skill set.} \textcolor{black}{An interesting open question in the FL setting is how MoPs affect existing privacy guarantees.}

% \vspace{-0.1cm}
% FL is often limited by communication/computation constraints, so model compression methods such as pruning and quantization are combined. We test MoP combining \texttt{Int8} quantization with different pruning ratios in Appendix \ref{sec:app_quant}.

\textbf{Evaluating MoPs against LoRA and AI$^3$. } 
Even though current PEFT approaches such as LoRA \citep{hu2021lora} and AI$^3$ \citep{liu2022fewshot} may not be completely compatible in terms of setup, it is still beneficial to compare against them empirically.
%The key difference in MoPs is that soft prompts are limited to modifying the input to the model. %\footnote{Even if we claim to inject prompts at an intermediate layer, MoPs can still be used in scenarios where we start from a pretrained model and the ``intermediate layer'' becomes the input layer.}. 
%Previous studies \cite{wang2023universality} have shown that soft-prompts are not as effective as modifying the weights of the model, which is the case of LoRA and AI$^3$, thus potentially limiting the maximum accuracy one can achieve. 
%MoPs is a non-invasive technique, as it edits information at the input layer, while other adapter methods require access to the (pretrained) model's parameters and architecture to be applied.
Previous studies \cite{wang2023universality} have shown that soft-prompts are not as effective as modifying the weights of the model, such as LoRA and AI$^3$, thus potentially limiting the maximum accuracy one can achieve. To overcome this limitation, we conducted experiments to demonstrate how we can orchestrate the different elements of MoPs --injection layer, number of prompts, number of experts-- to boost the expert skills and match similar performance as LoRA and AI$^3$ under different pruning ratios. 
Table \ref{tab:uncompressed_results} reports final perplexities on Prompt-Tuning (Baseline), MoPs, LoRA, and AI$^3$ after 20k training steps in centralized training.
These indicate that variations of MoPs can achieve comparable performance to the state-of-the-art models, overcoming the limitations of using soft prompts.

%The last column shows the gain of MoPs compared to regular Prompt-Tuning: while the performance of Prompt-Tuning decreases as the pruning ratio increases, MoPs close the gap with LoRA and AI$^3$. 

\begin{table}[ht]
\vspace{-0.5em}
\caption{Final perplexities; the implementation details for LoRA and AI$^3$ are described in Appendix E.}
\label{tab:uncompressed_results}
\vspace{-.7em}
\begin{center}
\begin{small}
\resizebox{1\columnwidth}{!}{
\begin{tabular}{l|c|c|c|c|c}
    \toprule
    Compression	& Baseline & MoPs & LoRA & AI$^3$ & MoPs Gain \\
    \midrule\midrule
    Uncompressed & 3.24 & \textbf{3.09} & 3.14 & 3.11 & \textbf{\textcolor{teal}{+0.15   (5\%)}}\\
    \midrule
    \textit{Trainable Params} & \multicolumn{5}{c}{0.01\%} \\
    \midrule\midrule
    75\% Pruned & 8.25 & \textbf{6.0} & 8.7 & 6.2 & \textbf{\textcolor{teal}{+2.25   (38\%)}}\\
    85\% Pruned & 18.16 & \textbf{12.4} & \textbf{12.4} & \textbf{12.4} & \textbf{\textcolor{teal}{+5.76   (46\%)}}\\
    90\% Pruned & 52.65 & 26.6 & \textbf{21.4} & 26.2 & \textbf{\textcolor{teal}{+26.05   (98\%)}}\\
    % \midrule
    % \textit{Trainable Params} & 0.01\% & 0.01\% & 0.01\% \\
    %\midrule\midrule
    %\texttt{Int8} & 8.9 & 7.4 & \textbf{6.6} \\
    %\texttt{Int8} + 75\% & 28.2 & \textbf{14.6} & 15.1 \\
    %\texttt{Int8} + 85\% & 70.9 & \textbf{22.5} & 26.0 \\
    %\texttt{Int8} + 90\% & 140.8 & 39.3 & \textbf{36.3} \\
    %\midrule
    %\textit{Trainable Params} & 0.01\% & 0.01\% & 0.01\% \\
    \bottomrule\bottomrule
    \end{tabular}}
\vspace{-1em}
\end{small}
\end{center}
\end{table}
 
\textbf{Appendix material.} 
Appendix A contains information how the federated data distribution is generated; Appendices B and C have experiments on how the gating function performs in centralized and federated learning settings, respectively; Appendix D contains experiments where MoP is combined with quantization techniques. Appendix F provides examples of \texttt{Int8} quantization with different pruning ratios in the centralized learning setup, where the number of prompts is increased to improve the granularity of the experts.
Appendix G includes results using the Phi-2 model \cite{li2023textbooks} as a baseline.
%Appendix \ref{sec:app_peft} contains information about the settings in the comparison with other PEFT methods.

\section{Conclusions}
\label{sec:discussion}

% MoPs allow the identification of relevant skills for the current task and dynamically select and combine prompts accordingly. This overcomes prompt training interference from multi-tasks across centralized and federated learning scenarios. 
% The results suggest that the gating function helps to overcome model drift problems resulting from heterogeneous data distribution in multi-source (federated) learning scenarios. 
% This is achieved by locally weighing and relatively updating the prompts for local data. 
% %With no additional cost, MoPs provide a powerful tool for overcoming interference from recovery of different skills from model compression, by embedding such skills in separated prompts. Overall, 
% Overall, MoPs method is a promising approach for improving the efficiency and effectiveness of prompt-based learning systems. 

MoPs allow the identification of relevant skills for the current task and dynamically select and combine prompts accordingly. This overcomes prompt training interference from multi-tasks across centralized and federated learning scenarios. Further results show how the gating functionality boosts soft-prompts to match similar performance as other state-of-the-art PEFT methods. 
The results suggest that the gating function helps to overcome model drift problems resulting from heterogeneous data distribution in multi-source (federated) learning scenarios. %This is achieved by locally weighing and relatively updating the prompts for local data.

\clearpage

\bibliography{aaai25}

\clearpage

\appendix
\onecolumn

\section{A. Federated skew distribution}
\label{sec:fl_data_dist}

To simulate a highly skewed data distribution in the across the clients for the federated learning experiments, we randomly selected total 5000 samples from all task categories. To simulate task and data heterogeneity, for data from each task category, we further split them into N partitions with different number of data samples (where N is the number of clients). To simulate the extreme data heterogeneity in real life scenario, we make one of the partition to have most of the data (it contains 15 times more samples than the rest partitions). We then randomly assigned one partition from each category to each client, resulting in different proportions and sizes of mixed tasks across the clients. %\rsim{would a figure help here?}

\section{B. Centralized Training - Gating function Analysis}
\label{sec:app_cl_gate_analysis}

\begin{figure}[h]
\begin{center}
\centerline{\includegraphics[width=0.8\columnwidth]{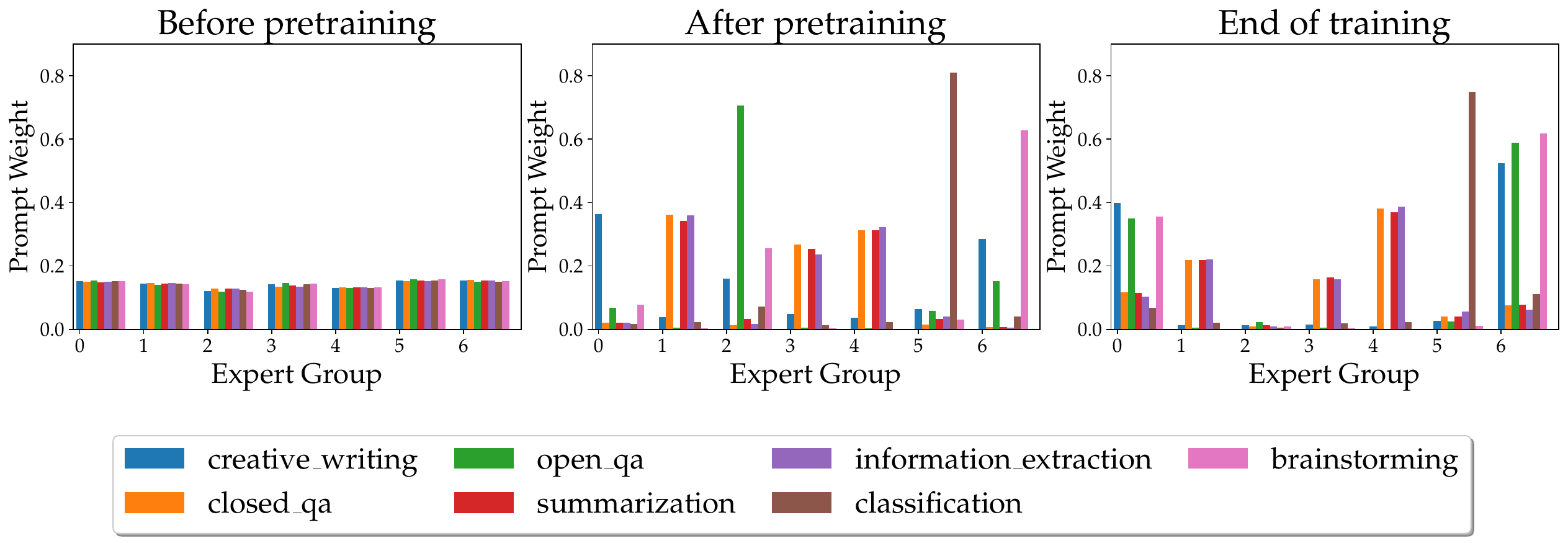}}
\caption{Averaged Prompt weight assigned each prompt group by gating function for test dataset using 85\% unstructured pruning Llama-7B in centralized setup.}
\label{fig:cl_gate_analysis}
\vspace{-2em}
\end{center}
\end{figure}

We further analyze how our gating function performs the assignment depending on the current task. 
In Figure \ref{fig:cl_gate_analysis}, we observe that the pretraining step helps the gating function to roughly distinguish between data domains/tasks, by encouraging one-to-one relationship between prompt experts and data domains/tasks. 
After training is done, instead of one-to-one relationship between prompt experts and data domains/tasks, we can see that our gating function learns to select the same expert group of prompts for similar tasks.
This suggests that our gating function has learned to adjust the prompt weight distribution, in order to better capture the domain/task relationship and specialize the expert assignment. 

%\vspace{-1em}

Below, we present the different results of the averaged prompt weights assigned to each prompt group by the gating function before, during, and after training steps for the Dolly-15k dataset in the structured/unstructured pruning. Different pruning ratios are displayed to demonstrate that more aggressive pruning ratios provide greater potential for improvement using the MoPs method.

\begin{figure}[H]
\begin{center}
\vspace{-1.5em}
\includegraphics[width=0.8\textwidth]
    {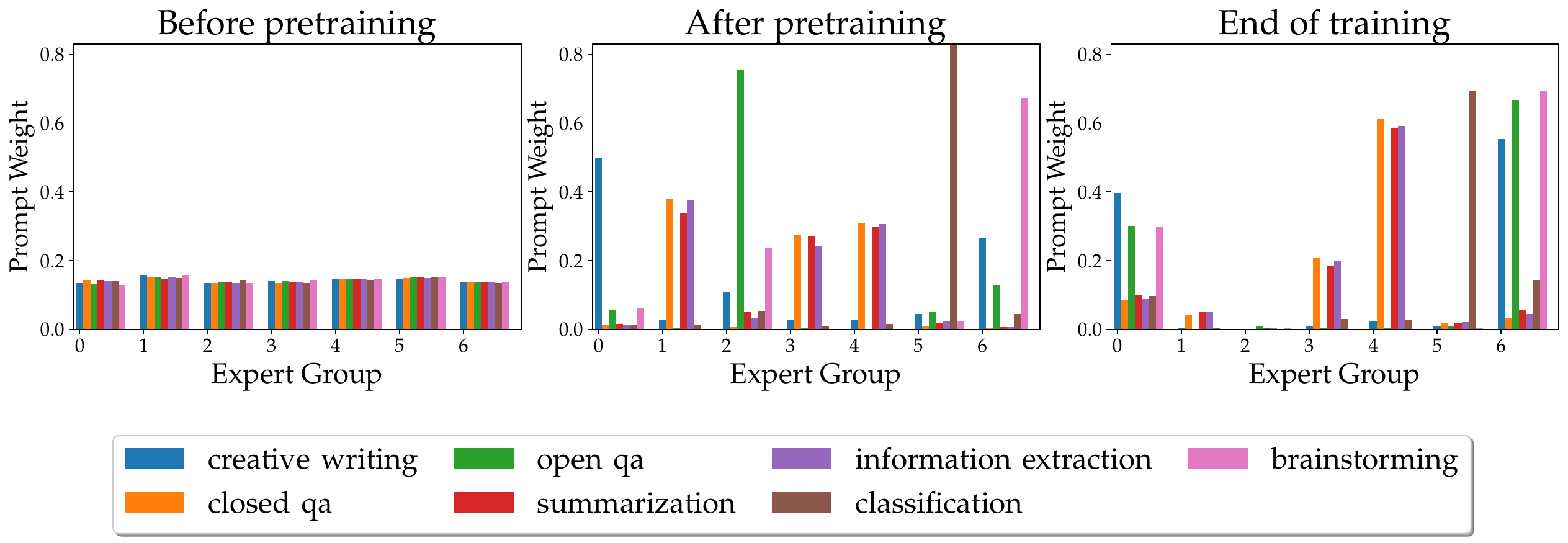}
    \caption{Averaged Prompt weight assigned each prompt group by gating function for test dataset using 75\% unstructured pruning Llama-7B}
    \label{fig:cl_gate_analysis_75}
\end{center}
\end{figure}

% \begin{figure}[h]
% \begin{center}
% \vspace{-1em}
% \includegraphics[width=0.7\textwidth]
%     {ICLR_2024/images/cl_analysis_90.pdf}
%     \caption{Averaged Prompt weight assigned each prompt group by gating function for test dataset using 90\% unstructured pruning Llama-7B}
%     \label{fig:cl_gate_analysis_90}
% \end{center}
% \end{figure}

\begin{figure}[H]
\begin{center}
%\vspace{-1.5em}
\includegraphics[width=0.8\textwidth]
{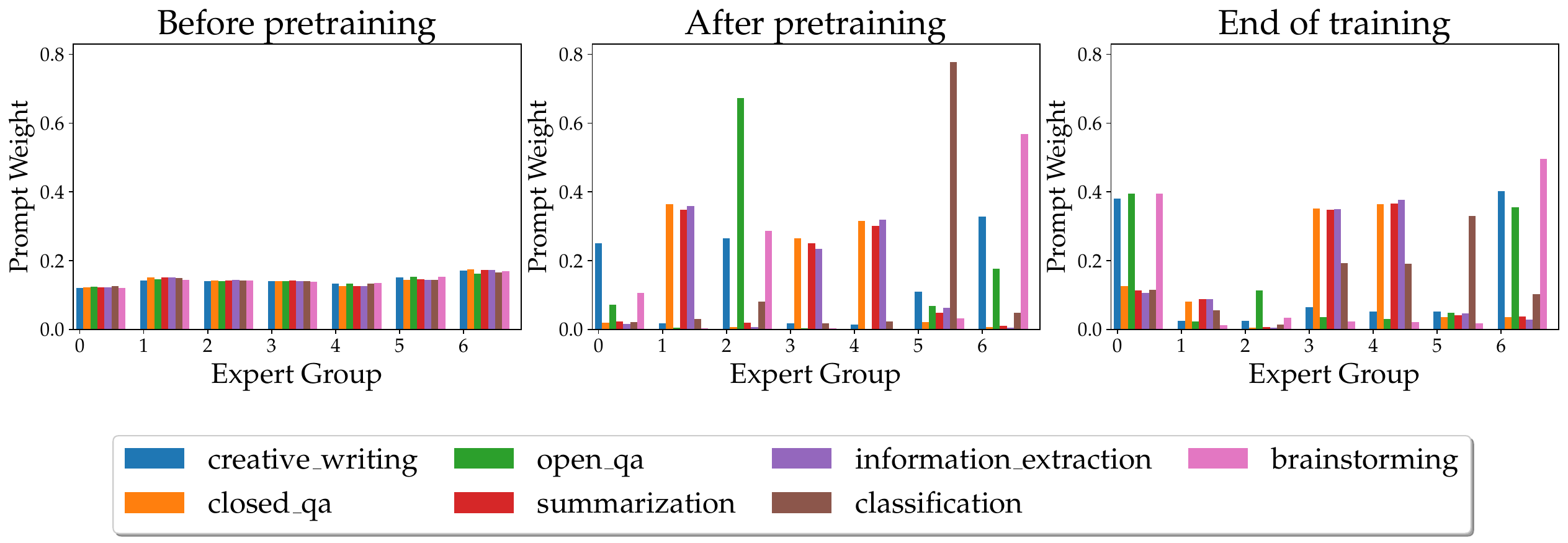}
    \caption{Averaged Prompt weight assigned each prompt group by gating function for test dataset using 7:8 (50\%) structured pruning Llama-7B}
    \label{fig:cl_gate_analysis_875_7_8}
\end{center}
\end{figure}

% \begin{figure}[h]
% \begin{center}
% \includegraphics[width=0.7\textwidth]
%     {ICLR_2024/images/cl_analysis_75_3_4.pdf}
%     \caption{Averaged Prompt weight assigned each prompt group by gating function for test dataset using 3:4 (75\%) structured pruning Llama-7B}
%     \label{fig:cl_gate_analysis_75_3_4}
% \end{center}
% \end{figure}

% \begin{figure}[h]
% \begin{center}
% \includegraphics[width=0.7\textwidth]
%     {ICLR_2024/images/cl_analysis_5_2_4.pdf}
%     \caption{Averaged Prompt weight assigned each prompt group by gating function for test dataset using 2:4 (50\%) structured pruning Llama-7B}
%     \label{fig:cl_gate_analysis_5_2_4}
% \end{center}
% \end{figure}

% \begin{figure}[h]
% \begin{center}
% \includegraphics[width=0.7\textwidth]
%     {ICLR_2024/images/cl_analysis_5_4_8.pdf}
%     \caption{Averaged Prompt weight assigned each prompt group by gating function for test dataset using 4:8 (50\%) structured pruning Llama-7B}
%     \label{fig:cl_gate_analysis_5_4_8}
% \end{center}
% \end{figure}

\section{C. Federated Training - Gating function Analysis}
\label{sec:app_fl_gate_analysis}

%Results on dolly-15k dataset

Similarly to the previous section, we show additional advantages provided by our method in the federated scenario. The alignment of the updates on the different experts helps minimize the effect of task interference. 

% \begin{figure}[htp] 
%     %\vspace{-0.3cm}
%     \includegraphics[width=1\textwidth]
%     {ICLR_2024/images/fl_analysis_75.pdf}
%     \vspace{-0.6cm}
%     \caption{Averaged Prompt weight assigned each prompt group by gating function for test dataset using 75\% unstructured pruning Llama-7B}
%     \label{fig:fl_gate_analysis_75}
%     \vspace{-0.2cm}
% \end{figure}

\begin{figure}[H]
\begin{center}
%\vspace{-1.5em}
\includegraphics[width=0.8\textwidth]
    {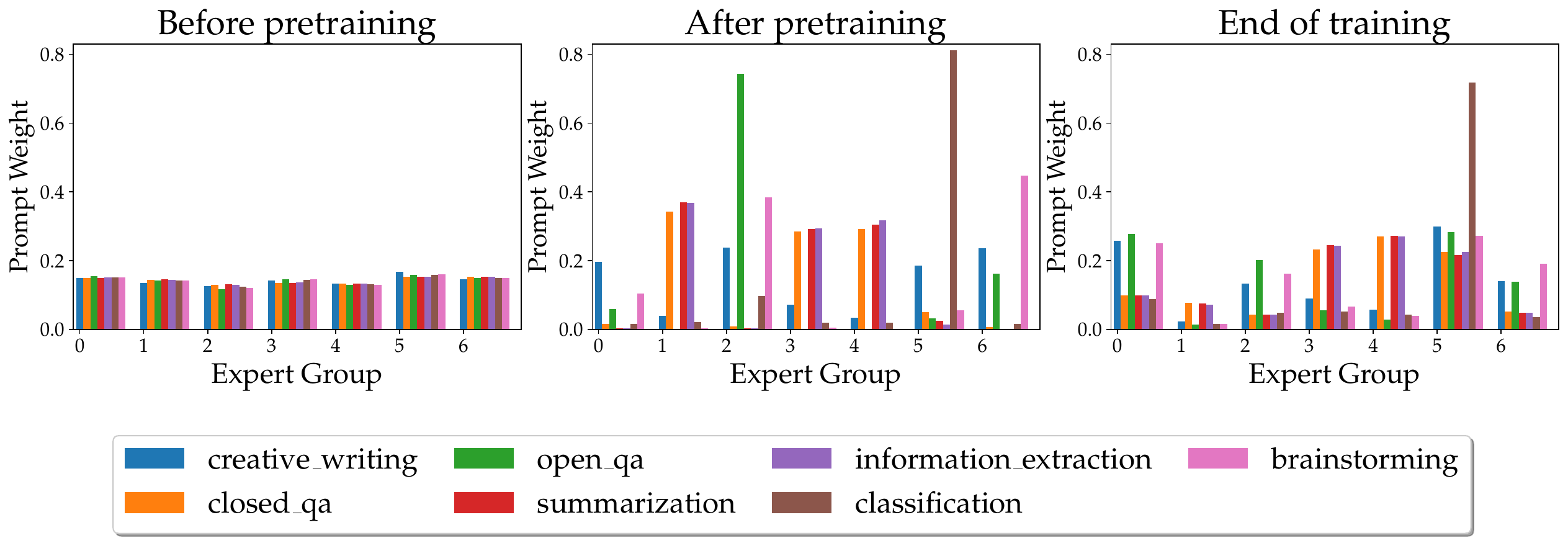}
    \caption{Averaged Prompt weight assigned each prompt group by gating function for test dataset using 85\% unstructured pruning Llama-7B}
    \label{fig:fl_gate_analysis_85}
\end{center}
\end{figure}

% \begin{figure}[htp] 
%     %\vspace{-0.3cm}
%     \includegraphics[width=1\textwidth]
%     {ICLR_2024/images/fl_analysis_90.pdf}
%     \vspace{-0.6cm}
%     \caption{Averaged Prompt weight assigned each prompt group by gating function for test dataset using 90\% unstructured pruning Llama-7B}
%     \label{fig:fl_gate_analysis_90}
%     \vspace{-0.2cm}
% \end{figure}

\begin{figure}[H]
\begin{center}
\includegraphics[width=0.8\textwidth]
{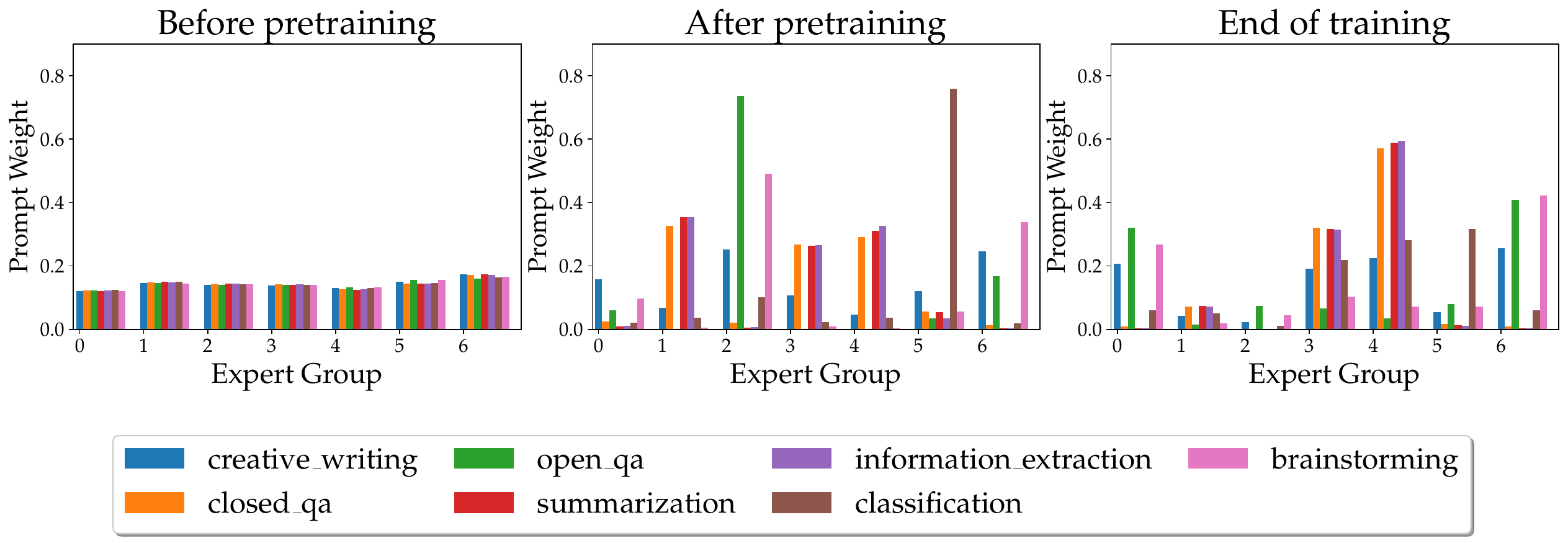}
    \caption{Averaged Prompt weight assigned each prompt group by gating function for test dataset using 7:8 (50\%) structured pruning Llama-7B}
    \label{fig:fl_gate_analysis_875_7_8}
\end{center}
\end{figure}

\begin{figure}[H]
\begin{center}
\includegraphics[width=0.8\textwidth]
{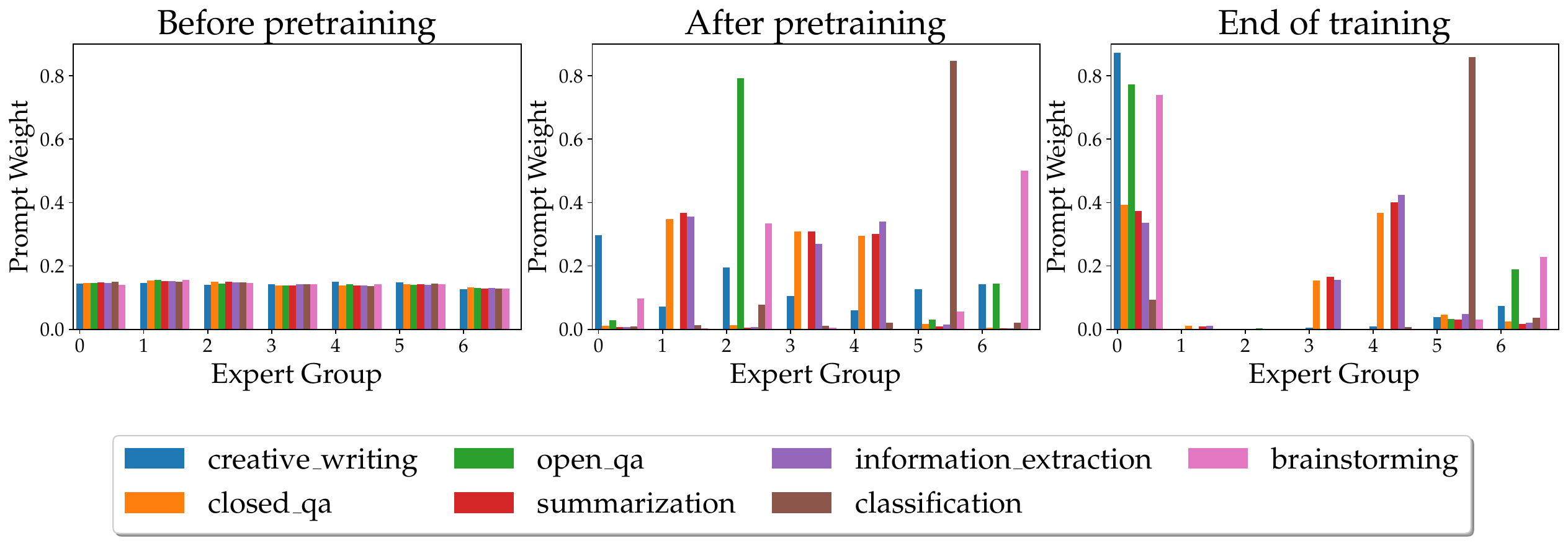}
    \caption{Averaged Prompt weight assigned each prompt group by gating function for test dataset using 2:4 (50\%) structured pruning Llama-7B}
    \label{fig:fl_gate_analysis_5_2_4}
\end{center}
\end{figure}

% \begin{figure}[htp] 
%     %\vspace{-0.3cm}
%     \includegraphics[width=1\textwidth]
%     {ICLR_2024/images/fl_analysis_2_4_0_5.pdf}
%     \vspace{-0.6cm}
%     \caption{Averaged Prompt weight assigned each prompt group by gating function for test dataset using 2:4 (50\%) structured pruning Llama-7B}
%     \label{fig:fl_gate_analysis_5_2_4}
%     \vspace{-0.2cm}
% \end{figure}

% \begin{figure}[htp] 
%     %\vspace{-0.3cm}
%     \includegraphics[width=1\textwidth]
%     {ICLR_2024/images/fl_analysis_4_8_0_5.pdf}
%     \vspace{-0.6cm}
%     \caption{Averaged Prompt weight assigned each prompt group by gating function for test dataset using 4:8 (50\%) structured pruning Llama-7B}
%     \label{fig:fl_gate_analysis_5_4_8}
%     \vspace{-0.2cm}
% \end{figure}

\section{D. Quantization Results}
\label{sec:app_quant}

To test MoP, we combined \texttt{Int8} quantization with different pruning ratios in FL. As seen in Table \ref{tab:quant_results}, MoPs outperformed the baseline in all cases but two case. 
MoP achieved the best results with medium pruning ratio. 
This result suggests that the effectiveness of a gating network can be significantly impacted by the pruning ratio. 
If the pruning ratio is too aggressive, the gating network will be rendered ineffective due to the poor embedding network. 
On the other hand, if the pruning ratio is too low, there may not be enough room for improvement compared to the baseline.

\begin{table}[h]
%\vspace{-1em}
\caption{\texttt{Int8} quantization with structured/unstructured pruning results on Dolly-15 dataset in the federated learning scenario with 10 clients.}
\label{tab:quant_results}
\vspace{-1em}
\begin{center}
\begin{small}
\resizebox{0.6\columnwidth}{!}{
\begin{tabular}{c|c|c|c|c|c}
    \toprule
    Dataset & Pruning method & Ratio & Baseline & MoP & Gain $\pm$ \\
    \midrule\midrule
	& Unstructured & \texttt{Int8}+90\% & 146.24 &  140.05 & \textbf{\textcolor{teal}{+6.19  ( 4\%)}} \\
	& Unstructured & \texttt{Int8}+85\% &  78.62 &  71.25 & \textbf{\textcolor{teal}{+7.37  ( 10\%)}} \\
	& Unstructured & \texttt{Int8}+75\% &  28.95 & 28.26 & \textbf{\textcolor{teal}{+0.69  ( 2\%)}}  \\
	\cmidrule(){2-6}
    Dolly-15k & Structured & \texttt{Int8}+7:8 (87.5\%) & 192.10 &  166.48 & \textbf{\textcolor{teal}{+25.62   (15\%)}}\\
    & Structured & \texttt{Int8}+3:4 (75.0\%) & 50.30 & 47.37 &  \textbf{\textcolor{teal}{+2.93  ( 6\%)}} \\
    & Structured & \texttt{Int8}+2:4 (50.0\%) & 14.24 & 14.51 &  -0.69  ( 2\%)\\
    & Structured & \texttt{Int8}+4:8 (50.0\%) & 13.13 & 13.10 & +0.03  ( 0\%)\\
    \bottomrule
    \end{tabular}}
\vspace{-1em}
\end{small}
\end{center}
\end{table}

\section{E. Evaluating MoPs performance against PEFT methods}
\label{sec:app_peft}

For the evaluation results in Table \ref{tab:uncompressed_results}, we used the Huggingface PEFT Hub \cite{peft} to evaluate the Llama-7B model on Dolly-15k dataset under the following setup:

\begin{itemize}
        \item Uncompressed Model 
        \item 75\% Unstructured Pruning
        \item 85\% Unstructured Pruning
        \item 90\% Unstructured Pruning    
\end{itemize}

The target modules specified for LoRA and AI$_3$ were chosen to ensure the same number of trainable parameters for each method. For the 90\% case we changed the layer injection in MoPs $L_\text{int}=1$, so we could reach the maximum performance on 70 prompts during training.

\begin{itemize}
  \item LoRA: 
    \begin{itemize}
    \vspace{-0.5em}
        \item lr: 1e-3
        \item r$:$2
        \item lora\textunderscore alpha: 32
        \item lora\textunderscore dropout: 0.05
        \item target\textunderscore modules: q\textunderscore proj,v\textunderscore proj
    \end{itemize}
    \vspace{-0.5em}
  \item AI$^3$:
    \begin{itemize}
        \vspace{-0.5em}
        \item lr: 1e-3
        \item target\textunderscore modules: q\textunderscore proj,v\textunderscore proj, q\textunderscore proj, gate\textunderscore proj
    \end{itemize}
  \item MoPs:
  \begin{itemize}
        \vspace{-0.5em}
        \item lr: 1e-3
        \item injection\textunderscore layer: 3
        \item prompts\textunderscore per \textunderscore expert: 10
        \item experts : 7
    \end{itemize}
\end{itemize}

\section{F. Pushing soft-prompts performance through MoPs "hyperparameters"}
\label{sec:app_quant_prun}

Table \ref{tab:quant_prun} shows the result of \texttt{Int8} + different pruning radios in the centralized setup. The last column indicates the relative gain of MoPs comparared with Prompt-Tuning (Baseline). We can observe that even when MoPs is injected in the 1st layer, the soft-prompts have a ceiling after these levels of compression in the original LLM. Table \ref{tab:quant_prun_setup} shows the setup in MoPs presented for each scenario.

\begin{table}[ht]
\caption{Summary of final perplexities reported on MoPs, LoRA, and AI$^3$ after 20k training steps in the centralized training setup. The implementation details for LoRA and AI$^3$ are described in Appendix E.}
\label{tab:quant_prun}
\begin{center}
\begin{small}
\resizebox{0.5\columnwidth}{!}{
\begin{tabular}{l|c|c|c|c|c}
    \toprule
    Compression	& Baseline & MoPs & LoRA & AI$^3$ & MoPs Gain \\
    \midrule\midrule
    \texttt{Int8} & 14.9 & 8.9 & 7.4 & \textbf{6.6} & \textbf{\textcolor{teal}{+6.01   (68\%)}}\\
    \midrule
    \textit{Trainable Params} & \multicolumn{5}{c}{0.01\%} \\
    \midrule
    \texttt{Int8} + 75\% & 40.3 & 28.2 & \textbf{14.6} & 15.1 & \textbf{\textcolor{teal}{+12.16   (43\%)}}\\
    \texttt{Int8} + 85\% & 95.6 & 70.9 & \textbf{22.5} & 26.0 & \textbf{\textcolor{teal}{+24.76   (35\%)}}\\
    \texttt{Int8} + 90\% & 197.7 & 140.8 & 39.3 & \textbf{36.3} & \textbf{\textcolor{teal}{+56.94   (40\%)}}\\
    \bottomrule\bottomrule
    \end{tabular}}
\vspace{-1em}
\end{small}
\end{center}
\end{table}

\begin{table}[ht]
\caption{MoPs "hyperparameters" settings for results presented in Table \ref{tab:quant_prun}}
\label{tab:quant_prun_setup}
\begin{center}
\begin{small}
\resizebox{0.5\columnwidth}{!}{
\begin{tabular}{l|c|c|c}
    \toprule
    Compression	& \# Experts & \# Prompts & Injection Layer \\
    \midrule\midrule
    \texttt{Int8}& 7 & 119 & 1 \\
    \texttt{Int8}+ 75\% & 7 & 119 & 1 \\
    \texttt{Int8}+ 85\% & 7 & 119 & 1  \\
    \texttt{Int8}+ 90\% & 7 & 119 & 1 \\
    \bottomrule\bottomrule
    \end{tabular}}
\end{small}
\end{center}
\end{table}

\section{G. Using the Phi-2 model as an alternative LLM basis.}
\label{sec:phi2}

In this section, we consider the Phi-2 model \cite{li2023textbooks}, particularly the Phi-2 - 2.7B parameters (Huggingface checkpoint).
We have implemented the injection of prompts in Layer 10 for all MoPs experiments using 7 experts with 10 prompts each (70 experts total).
We have removed the frozen prompts in layer 0 (since they are used for initial recovery from model compression, and here we are using the model without compression).
We are only relying on the middle layer prompts injected with the gating function.
The training includes 20K steps with a learning rate $10^{-4}$ on the Dolly-15k dataset.
 
For LoRA \cite{hu2021lora}, we are using rank size 32 on the following projections of the model ($Q$,$K$,$V$,$O$).
We use the same training data.
For MoPs + Pretrained LoRA, we use the final model produced by LoRA fine-tuning (using the best result) and then MoPs for another 20K steps (using the same setup above) to reach the final performance shown in the table below.

Table \ref{tab:phi2} depicts results for this ``small'' Language Model (SLM) case. 
The MoPs approach maintains its advantage. 
Notably, the combination of a pretrained LoRA model, fine-tuned on MoPs, yielded encouraging results, surpassing both MoPs and LoRA independently by an impressive 21\% while reducing memory footprint. 

\begin{table}[ht]
\caption{\textcolor{black}{Summary of relative gains over Pretrained LoRA Adapter using the Phi-2 model \cite{li2023textbooks}. Final PPL after 20k steps on Dolly-15K Dataset using Phi-2 Model. Reserved memory estimation using 4 bits precision. The reserved memory is obtained by multiplying the total number of model parameters with the bit precision (4 bits), divided by (1/1024) to obtain the minimum reserved memory needed to run inference. Note that this is for a single LoRA module, using more LoRA adapters will linearly increase the memory fragmentation.}}
\label{tab:phi2}
\begin{center}
\begin{small}
\resizebox{0.5\columnwidth}{!}{
\begin{tabular}{l|c|c|c|c}
    \toprule
    Method	& LoRA & MoPs & MoPs + Pretrained LoRA & MoPs Gain \\
    \midrule\midrule
    \texttt{Perplexity} & 37.9 & 36.62 & 31.28 & \textbf{\textcolor{teal}{+6.65   (21\%)}}\\
    \midrule
    \textit{Memory (GB)} & 10.41 & 10.36 & 10.36 & \\
    \bottomrule\bottomrule
    \end{tabular}}
\vspace{-1em}
\end{small}
\end{center}
\end{table}

\section{H. Hyperparameter settings for results presented in main text}

Table \ref{tab:uncompressed_setup} shows the setup in MoPs for each scenario in Table \ref{tab:uncompressed_results}. For the uncompressed case, MoPs are able to outperform LoRA and AI$^3$ using the same setup as Tables
\ref{tab:cl_results}-\ref{tab:fl_results}. On the other hand, for the pruning cases, we inject the gating function at earlier layers to match the performance of LoRA and AI$^3$, incurring additional back-propagation costs. %, so we can allow the experts to have more influence on the LLM and match similar numbers as .

\begin{table}[ht]
\vspace{-1.5em}
\caption{Table \ref{tab:uncompressed_results} MoPs "hyperparameter" settings.}
\label{tab:uncompressed_setup}
\begin{center}
\begin{small}
\resizebox{0.5\columnwidth}{!}{
\begin{tabular}{l|c|c|c}
    \toprule
    Compression	& \# Experts & \# Prompts & Injection Layer \\
    \midrule\midrule
    Uncompressed & 7 & 70 & 10 \\
    75\% Pruned & 7 & 70 & 3 \\
    85\% Pruned & 7 & 70 & 3  \\
    90\% Pruned & 7 & 70 & 1  \\
    \bottomrule\bottomrule
    \end{tabular}}
\vspace{-1em}
\end{small}
\end{center}
\end{table}

We observe that the gating functionality is able to leverage soft-prompt-based methods and achieve at least comparable performance with LoRA and AI$^3$, without being invasive. 
%This demonstrates the significant boost that our gating functionality can provide. 

\end{document}